\documentclass[letterpaper, 10 pt, journal, twoside]{IEEEtran}

\IEEEoverridecommandlockouts                              
\usepackage{graphics} 
\usepackage{tensor}
\usepackage{epsfig} 
\usepackage{times} 
\usepackage{amsmath} 
\usepackage{amssymb}  
\usepackage[hidelinks]{hyperref} 
\usepackage[table]{xcolor} 
\usepackage{arydshln} 
\usepackage{color} 
\usepackage{float}
\usepackage{multirow}
\usepackage{booktabs}
\usepackage{tikz}
\usepackage{wrapfig}
\usepackage{bm}
\usetikzlibrary{calc ,math, positioning, intersections, decorations, external, plotmarks}
\usepackage{pgfplots}
\pgfplotsset{compat=1.16}
\usepackage{subcaption}

\usepackage{tabularx}
\usepackage{mathtools}
\usepackage{enumerate} 
\usepackage{cite}
\usepackage{siunitx}
\usepackage{array}
\usepackage{tabularray}
\usepackage{algorithm}
\usepackage[noend]{algpseudocode}

\newcommand{\Rom}[1]{\uppercase\expandafter{\romannumeral #1\relax}}
\newcommand{\rom}[1]{\expandafter{\romannumeral #1\relax}}

\newcommand{\PreserveBackslash}[1]{\let\temp=\\#1\let\\=\temp}
\newcolumntype{C}[1]{>{\PreserveBackslash\centering}p{#1}}
\newcolumntype{R}[1]{>{\PreserveBackslash\raggedleft}p{#1}}
\newcolumntype{L}[1]{>{\PreserveBackslash\raggedright}p{#1}}

\newcommand{\fref}[1]{Fig. \ref{#1}}
\newcommand{\sref}[1]{Section \ref{#1}}
\newcommand{\tref}[1]{Table. \ref{#1}}
\newcommand{\iref}[1]{(\ref{#1})}
\newcommand{\argmin}{\operatornamewithlimits{argmin}}

\definecolor{roninpink}{HTML}{B970AF}
\definecolor{baselinegold}{HTML}{fdbb7d}
\definecolor{gtblue}{HTML}{0000FF}

\usepackage[colorinlistoftodos,prependcaption,textsize=tiny]{todonotes}
\title{\LARGE \bf AirIMU: Learning Uncertainty Propagation for Inertial Odometry}

\author{Yuheng Qiu$^{1}$, Chen Wang$^{2}$, Can Xu$^{1}$, Yutian  Chen$^{1}$, Xunfei Zhou$^{3}$, Youjie Xia$^{3}$, and Sebastian Scherer$^{1}$
\thanks{$^{1}$Robotics Institute, Carnegie Mellon University, Pittsburgh, PA 15213, USA. Email: {\tt\small \{yuhengq, canxu, yutianch, basti\} @andrew.cmu.edu}}%
\thanks{$^{2}$Spatial AI \& Robotics Lab, University at Buffalo, Buffalo, NY 14260, USA. Email: {\tt\small chenw@sairlab.org}}%
\thanks{$^{3}$OPPO US Research Center, Palo Alto, CA 94303, USA. Email: {\tt\small \{xunfei.zhou, youjie.xia\}@oppo.com}}%
}

\begin{document}
\makeatletter
\g@addto@macro\@maketitle{
  \captionsetup{type=figure}\setcounter{figure}{0}
  \def\mycolspace{1.2mm}
  \centering
    \includegraphics[width=\linewidth]{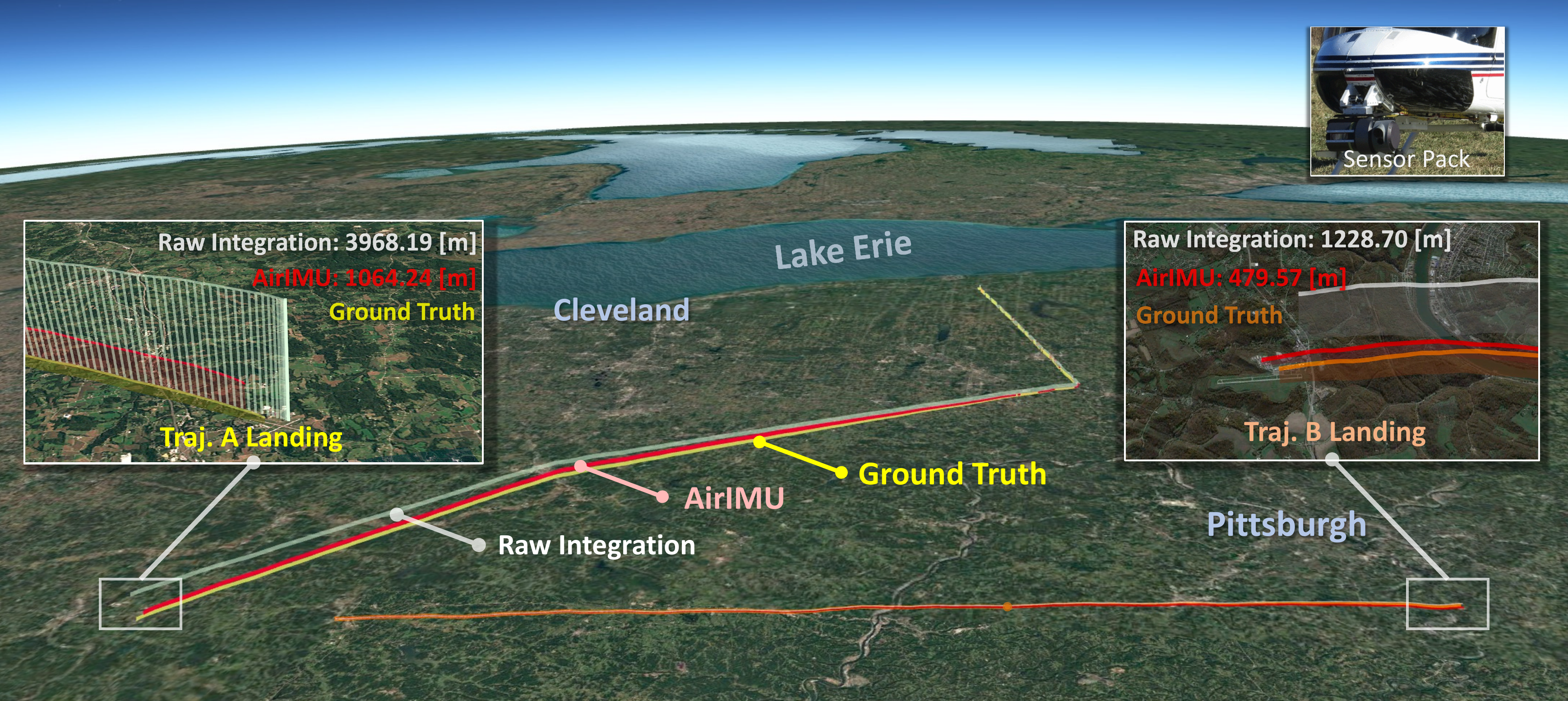}
	\captionof{figure}{AirIMU is a hybrid method that serves the dual purpose of state estimation and uncertainty estimation. AirIMU models the non-deterministic error and the corresponding uncertainty using a data-driven approach. Our model is adaptable to the full spectrum of IMUs, including high-end IMUs used in helicopter navigation. 
 }
	\label{fig:eyecatcher}
}
\makeatother
\maketitle
\begin{abstract}

Inertial odometry (IO) using strap-down inertial measurement units (IMUs) is critical in many robotic applications where precise orientation and position tracking are essential.
Prior kinematic motion model-based IO methods often use a simplified linearized IMU noise model and thus usually encounter difficulties in modeling non-deterministic errors arising from environmental disturbances and mechanical defects.
In contrast, data-driven IO methods struggle to accurately model the sensor motions, often leading to generalizability and interoperability issues.
To address these challenges, we present AirIMU, a hybrid approach to estimate the uncertainty, especially the non-deterministic errors, by data-driven methods and increase the generalization abilities using model-based methods.
We demonstrate the adaptability of AirIMU using a full spectrum of IMUs, from low-cost automotive grades to high-end navigation grades.
We also validate its effectiveness on various platforms, including hand-held devices, vehicles, and a helicopter that covers a trajectory of 262 kilometers.
In the ablation study, we validate the effectiveness of our learned uncertainty in an IMU-GPS pose graph optimization experiment, achieving a 31.6\% improvement in accuracy.
Experiments demonstrate that jointly training the IMU noise correction and uncertainty estimation synergistically benefits both tasks.
The source code and video are available at: \href{https://airimu.github.io}{https://airimu.github.io/}
\end{abstract}


\section{Introduction}
\label{Intro}
\IEEEPARstart{I}{nertial} Measurement Unit (IMU), capable of sensing high-frequency linear acceleration and angular velocity, is an essential component in almost all kinds of robots \cite{zhao2023subt, burri2016euroc, KITTIDataset}.
As a result, inertial odometry (IO), which aims to estimate the relative movements of robots using IMU measurements, has broad applications in autonomous driving \cite{KITTIDataset}, navigation \cite{qi2002direct}, augmented reality \cite{hincapie2015gyrowand}, and human motion tracking \cite{menolotto2020motion}.
Due to the IMUs' relative independence from the external environment, IO is insensitive to environmental disturbances such as illumination changes and visual obstructions \cite{zhao2023subt}. 
Therefore, IO has great potential in robust state estimation under complex operation conditions, especially compared to exteroceptive sensors like cameras and LiDAR \cite{qin2018vins}.
Nevertheless, the performance of existing IO algorithms, either the model-based \cite{forster2015imu} or the data-driven methods \cite{chen2018ionet}, is still unsatisfactory for real-world applications \cite{zhao2023subt}.

\begin{figure*}[!t]
	\centering
    \includegraphics[width=\linewidth]{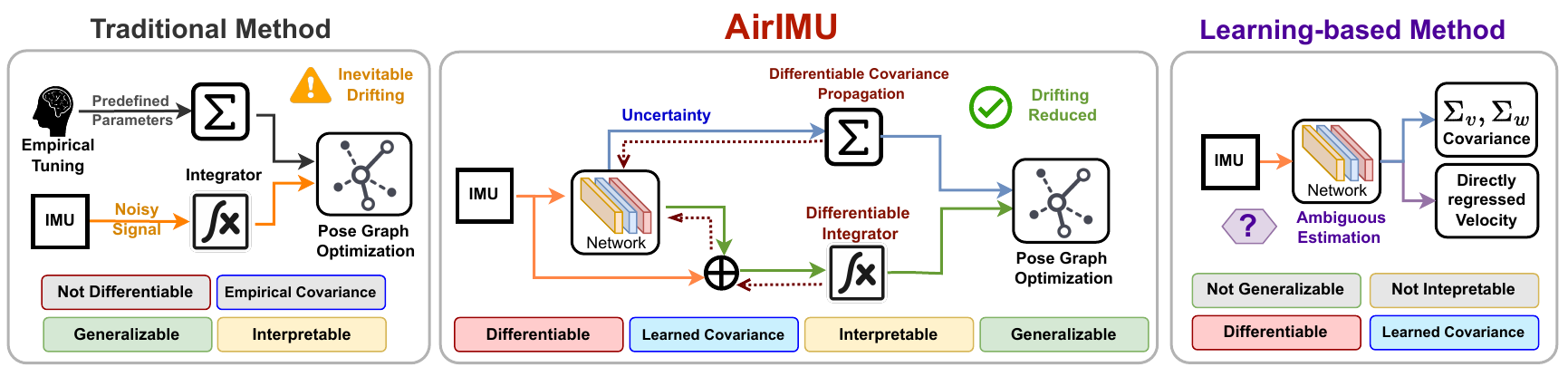}
    \caption{
        \textbf{Left:} Traditional methods with empirically set uncertainty parameters often fail to capture noise inherent in the sensor accurately.
        \textbf{Middle:} AirIMU corrects the IMU noise and predicts the corresponding uncertainty of the integration. Notably, its covariance propagation is differentiable, promoting efficient training.
        \textbf{Right:} Learning-based methods predict velocity and the corresponding covariance but lack interpretability and generalizability across different data domains.}
	\label{Fig:1}
\end{figure*}




Kinematic model-based IO methods have excellent generalizability but often suffer from complex and multifaceted noise \cite{barshan1995inertial}.
The IMU sensor noises can be classified into two categories: the deterministic errors and the non-deterministic errors \cite{rehder2016extending}.
The deterministic errors are the systematic biases in the measurements that can be consistently predicted or modeled, often arising from factors such as installation imperfections, and calibration errors \cite{li2022calib}.
Conversely, the non-deterministic errors, which are also known as stochastic errors, are related to factors like mechanical shocks, vibration influences, thermal effects, mechanical defects, and calculation ambiguities \cite{jao2023prio, capriglione2020experimental, li2014online}.
Many methods employ linearized models to handle the deterministic errors by accounting for additive sensor bias, scale factors, and axis misalignment \cite{tedaldi2014robust, furgale2013unified, Krebs2012} and handle non-deterministic errors using zero-mean Gaussian function \cite{qin2018vins, rehder2016extending}.
However, these simplified linear models struggle to fully account for non-deterministic uncertainties from the natural environments, highlighting a critical area for improvement. 
Data-driven IO, which use neural networks to predict the motion and implicitly fit the IMU noise, often face challenges with interpretability and generalizability.
To be more specific, these methods often overfit in specific motion patterns such as pedestrian \cite{herath2020ronin} or vehicle \cite{brossard2020ai}, and fail to generalize to scenarios outside of the training distribution.
For example, the model trained on hand-held platforms running on flat ground fails to generalize to robots claiming stairs or drones \cite{buchanan2022deep}.
Moreover, replacing the motion models with the neural networks undermines the model's interpretability.

Recent advances in hybrid methods show the potential to merge the benefits of both the model-based method and the data-driven method. 
For example, some researchers fuse the data-driven motion model with the IMU kinematic motion model using EKF \cite{liu2020tlio, sun2021idol} or batched optimization \cite{buchanan2022learning}.
These solutions improve the robustness of the state estimation but still struggle with generalizability.
On the other hand, efforts have been made in training neural networks to model the error in gyroscope \cite{brossard2020ai}, accelerometer\cite{zhang2021imu}, IMU bias evolution \cite{buchanan2022deep}.
Despite these advancements, none of these algorithms explicitly model the uncertainty of the IMU measurements in the IMU pre-integration.
When deploying the data-driven model with visual-inertial odometry \cite{zhang2021imu, buchanan2022deep}, these algorithms still assume the IMU covariance model as a manually calibrated stochastic variable.
These gaps hinder the deployment of hybrid methods for real-world robots.


To close these gaps, we present a hybrid method AirIMU as illustrated in \fref{Fig:2}. It leverages the data-driven method to model the non-deterministic noise and the kinematic motion model to ensure generalizability in novel environments. 
To empower the AirIMU with the ability to learn the uncertainty and noise model, we build a differentiable IMU integrator and a differentiable covariance propagator on manifolds.
Unlike the previous data-driven covariance prediction models \cite{liu2020tlio, sun2021idol} that only predict the covariance of the output translation within a fixed window, we predict the accumulated covariance for a long-time pre-integration.
Moreover, we jointly train the uncertainty module with the noise correction module, which can capture better IMU features and benefit both tasks.

In the experiments, we demonstrate the adaptability to a full spectrum of IMUs encompassing various grades and modalities, as illustrated in \fref{Fig:Devices}.
To validate the effectiveness of our learned covariance in sensor fusion, we design a pose graph optimization (PGO) system that fuses the AirIMU outputs with the GPS signals. 
To the best of our knowledge, we are the first to train a deep neural network that models IMU sensor uncertainty through differentiable covariance propagation.
In summary, the main contributions of this paper are:
\begin{itemize}

\item This paper proposes AirIMU, a hybrid method that leverages the data-driven method to model the non-deterministic error, and the kinematic motion model to ensure generalizability in novel environments. 

\item We propose to jointly train the IMU noise and uncertainty through the differentiable integrator and covariance propagator.
This training strategy improves 17.8\% of the IMU preintegration results in Subt-MRS benchmark \cite{zhao2023subt}.

\item Our method demonstrates adaptability across multiple grades of IMUs, from low-cost automotive-grade IMUs to high-end navigation-grade IMUs.
We further evaluate AirIMU on the large-scale visual terrain navigation dataset ATLO\cite{cisneros2022alto} cruising over 262 \si{\km}.
Additionally, we illustrate the effectiveness of the learned uncertainty in a GPS-IMU PGO experiment, achieving a 31.6\% improvement in optimization.

\item We develop a tool package for the differentiable IMU integrator and covariance propagator that supports batched operations and the cumulative product, which speeds up the training process.
This package is grounded on \textit{PyPose}\cite{wang2023pypose} library and the tutorial is now open-source at:  \href{https://pypose.org/tutorials/imu/imu_integrator_tutorial}{https://pypose.org/tutorials/imu}

\end{itemize}

\section{Related Works}

\subsection{Traditional Inertial Odometry}
\label{sec: imu-denoise}
Traditional model-based IO algorithms utilize the IMU kinematic motion model \cite{forster2015imu} to estimate the relative motion and are often integrated with other auxiliary sensors like GPS, Camera\cite{qin2018vins}, and LiDAR \cite{hemann2016long} with filter-based method or batched optimization.
To deploy IO on mobile robots, especially those equipped with low-cost IMUs, modeling IMU deterministic errors is indispensable \cite{barshan1995inertial}, which generally include the additive sensor bias, scale factors, and axis misalignment \cite{rehder2017camera}. 

To address the non-deterministic error, factors like nonlinearity, g-sensitivity\cite{Krebs2012}, size effect \cite{hung1979size}, vibration\cite{capriglione2020experimental}, thermal effect \cite{niu2013fast}, and Coriolis force \cite{brossard2021associating} are rigorously addressed.
These methods, however, rely on expensive motion simulators such as shakers and rate tables \cite{qureshi2017algorithm}, which is not available in practice.
To calibrate the low-cost IMUs, the IO algorithms utilize the IMU measurements under static conditions\cite{tedaldi2014robust, barshan1995inertial}, or auxiliary sensors like the camera\cite{rehder2016extending, yang2023online} and LiDAR\cite{lv2020targetless} for calibration.
Kalibr \cite{rehder2016extending} is one of the most widely used calibration algorithms, calibrating the IMU intrinsic and extrinsic parameters between the camera and the IMU.
However, all these calibration methods focus on modeling the linear behavior of the IMU sensor under static conditions or controlled environments, which often struggle to fully capture the noise inherent in the natural environment and sensor defects.
This reveals a significant gap in the current IMU calibration and the practices of IO.

\subsection{Learning Inertial Odometry}
\label{Sec: LIO}
A growing trend in leveraging a data-driven model for IO has been witnessed, such as IMU odometry network \cite{yan2018ridi, chen2018ionet}, EKF-fused IMU odometry \cite{liu2020tlio, sun2021idol, cao2022rio} and learning-based IMU calibration \cite{brossard2020denoising}.
In RIDI \cite{yan2018ridi}, a supervised Support Vector Machine (SVM) classifier is introduced to determine IMU placement modes (e.g., handheld, bag) using accelerometer and gyroscope measurements.
Mode-specific Support Vector Regressions (SVRs) are then employed to estimate pedestrian velocities based on motion types, enabling corrective adjustments to sensor readings.
IONet \cite{chen2018ionet} introduced deep learning to regress the translation within a fixed temporal window. 
Building on this concept, RoNIN \cite{herath2020ronin} improves the accuracy of the data-driven IO with multiple neural network architectures, including Temporal Convolutional Networks (TCN) and Long Short-Term Memory (LSTM).
RIO \cite{cao2022rio} explores rotation-equivalence as a self-supervisory mechanism for training IO models. 
These studies utilize data-driven motion estimation models to replace the IMU integration process, thereby reducing the drift associated with IMU double integration. 
Inspired by these IMU odometry network studies, TLIO \cite{liu2020tlio} integrates the data-driven motion estimation model with the IMU integrator through a stochastic-cloning Extended Kalman Filter (EKF). 
This integrated approach optimizes position, orientation, and sensor biases, solely with IMU data.
IDOL \cite{sun2021idol} further extends this framework with a data-driven orientation estimator, generating more accurate results with better consistency.
Although these methods are promising in pedestrian datasets like RoNIN \cite{herath2020ronin}, the motion estimation heavily relies on prior knowledge about human movements, such as the consistent walking pattern of pedestrians and constant velocity assumption, which fail to generalize to other platforms like drones and vehicles.
Moreover, these studies often assume the IMU measurement is transformed to the world coordinates or aligning the y-axis with gravity, which is impractical in real scenarios. 


Inspired by the advancements of the data-driven methods in modeling unknown parameters, prior researchers employ neural networks to implicitly calibrate the IMUs \cite{nobre2019learning, brossard2020denoising, zhang2021imu}. 
Brossard et al. \cite{brossard2020denoising} train a Convolutional Neural Network (CNN) to denoise the low-frequency error in gyroscopes.
Zhang et al. \cite{zhang2021imu} employed a Long Short-Term Memory (LSTM) network to preprocess the raw IMU, and subsequently channeled the preprocessed IMU measurements into a filter-based visual-inertial odometry system.
Deep IMU Bias \cite{buchanan2022deep}, utilizes neural networks to model the sensor bias, and incorporates the learned bias as an additional constraint in the bundle adjustment.
These studies showcase the potential of data-driven methods in modeling IMU errors, especially when the sensor's physical conditions are difficult to capture.
However, none of the referenced algorithms have explicitly tackled the challenge of modeling the uncertainty for IMU integration.
When deploying these data-driven models, the predefined covariance parameters can no longer fit the learned IMU sensor model, undermining the IO's accuracy and robustness. 




\begin{figure}[!t]
	\centering
    \includegraphics[width=0.95\linewidth]{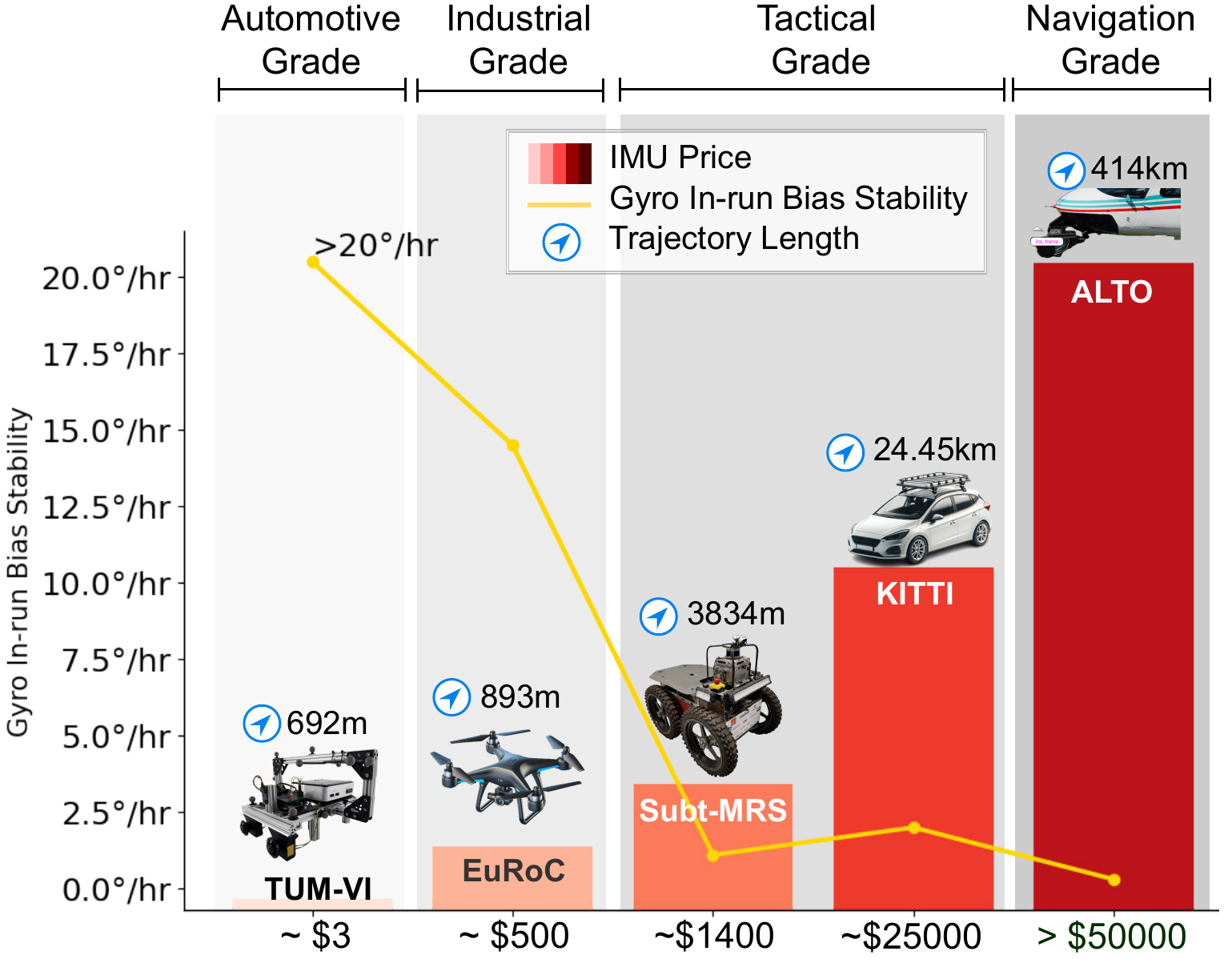}
    \caption{
        We evaluate our algorithm on IMUs with different prices and drift rates. Sensor quality is evaluated based on the gyroscope’s in-run bias stability metric. Our evaluation spans the full spectrum of IMUs, including automotive-grade, industrial-grade, tactical-grade, and navigation-grade IMUs.}
	\label{Fig:Devices}
\end{figure}

\subsection{IMU uncertainty model}
The uncertainty model in IO, which quantifies the reliability of inertial measurements, is the key to optimizing both filter performance and the batched optimization results. 
Traditional IO follows the kinematics model formulated in the IMU integration to propagate the covariance matrix from the predefined IMU uncertainty parameters \cite{forster2015imu}.
A commonly used solution to calibrate the uncertainty parameters is the Allen deviation \cite{el2007analysis}, which quantifies the sensor bias and bias instability by analyzing the variance of the sensor errors in a static position.
However, these parameters often require empirical tuning and manual re-calibration during testing due to unknown environmental disturbances and mechanical imperfections.

Recent researches in data-driven methods demonstrate the potential to model the uncertainty model with neural network \cite{russell2021multivariate}.
The AI-IMU \cite{brossard2020ai} proposes a dead-reckoning algorithm for vehicles, which assumes that the lateral and vertical velocities are approximately null.
To model the confidence degree of these null assumptions, the authors train a CNN network from IMU to predict two constant uncertainty parameters corresponding to lateral and vertical velocity.
DualProNet \cite{solodar2023vio} is another attempt to use a deep neural network to predict the uncertainty parameters in IMUs.
This work, however, requires high-frequency ground truth covariance labels for all IMU measurements, which are generally non-available during practice.
In TLIO \cite{liu2020tlio}, the authors jointly train a covariance model of the predicted velocity with a fixed temporal window, which is later fused in an EKF. 

In response to these challenges, our method leverages the differentiable covariance propagation to learn the high data-rate uncertainty. 
This approach alleviates the need for high-frequency ground truth labels and bridges the gap between uncertainty modeling and practical limitations.
\section{AirIMU Model}


To balance the accuracy and generalizability, the AirIMU model decouples the non-deterministic error model from the precisely defined kinematic motion model, as depicted in \fref{Fig:2}. 
Specifically, it utilizes a data-driven model to capture the error and uncertainty inherent in the IMU, while employing the kinematic-based method to integrate the state and covariance.
By supervising the integrated state and covariance, the AirIMU jointly trains the sensor error and uncertainty models throughout the differentiable integrator.
To accelerate the training speed with long sequences, we propose a tree-like structure for cumulative operation, and redesign the covariance propagation to facilitate batched operation and parallel computing.
The refined state and covariance outputs are subsequently incorporated into the PGO module, where they are fused with auxiliary sensors such as GPS.

Overall, we show the data-driven model in \sref{Model: Network}, the kinematic model in \sref{Model: Integrator}, loss functions in \sref{Model: Loss}.
Moreover, we detail our efforts of accelerating the integration and covariance propagation in \sref{Model: Accelerate}. Lastly, we present the PGO module that fuses the AirIMU with auxiliary sensors like GPS in \sref{Model: System}.

\subsection{Data-driven module}
\label{Model: Network}
\begin{figure}[!t]
	\centering
    \includegraphics[width=\linewidth]{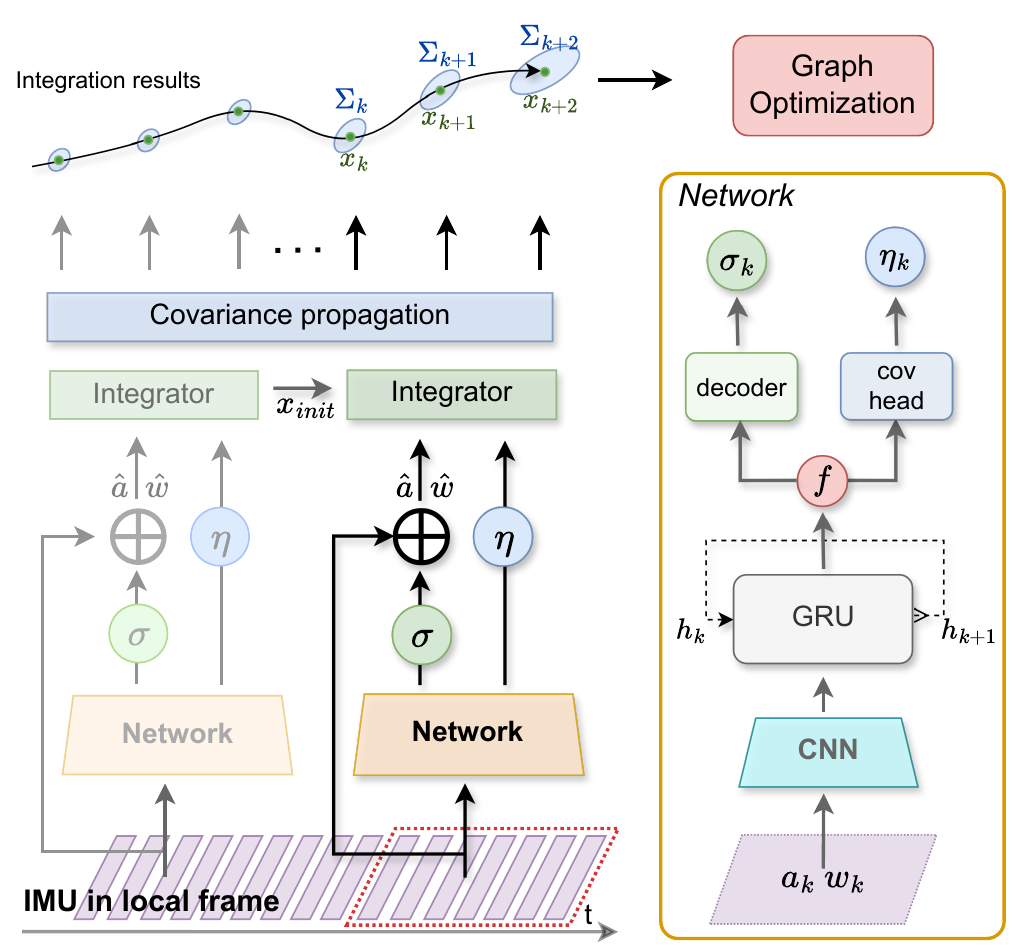}
    \caption{
    \textbf{Right:} We employ a CNN-GRU encoder to capture IMU features ($f$) from raw IMU accelerometer ($a_k$) and gyroscope ($w_k$) measurements.
    Subsequently, these features are decoded to obtain IMU corrections ($\sigma_{acc}$, $\sigma_{gyro}$).
    \textbf{Left:} We add the raw measurements with learned corrections, generating corrected IMU measurements ($\hat a_k$ and $\hat w_k$). The learned uncertainties ($\eta$) are propagated to estimate the corresponding covariance matrix ($\Sigma$).
}
	\label{Fig:2}
\end{figure}

As shown in the right part of \fref{Fig:2}, the AirIMU model utilizes an encoder-decoder network structure to process raw IMU data, which includes three-axis acceleration $a$ and three-axis angular velocity $w$, both measured in the body frame ${B}$. The network outputs corrections for acceleration and angular velocity, denoted as ${}^B\sigma^{acc}$ and ${}^B\sigma^{gyro}$, respectively. Additionally, it calculates the corresponding uncertainty measures for the gyroscope $\eta^{gyro}$ and accelerometer $\eta^{acc}$.

To leverage and understand the hidden relationship between the IMU noise and uncertainty, we train a shared encoder network to capture the common representations of the raw IMU measurements for both tasks.
In the encoder model, we first utilize a 1D CNN network to learn the lower-level features like the pattern of the accelerometer and gyroscope.
To capture the temporal dependency from the previous IMU data, we combine the CNN network with gated recurrent unit (GRU) layers to generate the IMU feature.

With the shared features, we build a correction decoder to estimate the error ${}^B{\sigma^{gyro}}, {}^B{\sigma^{acc}}$ in the following:
\begin{equation}
    \left[{}^B{\sigma^{gyro}}, {}^B{\sigma^{acc}}\right] = f_\pi\left({}^Bw, {}^Ba\right),
\end{equation}
where $\pi$ represents the learnable parameters of the decoder. 
For the covariance, we train another decoder with a similar structure to predict the uncertainty of each corresponding frame. 
This uncertainty is predicted by a covariance model denoted as:
\begin{equation}
    \left[\eta^{gyro}, \eta^{acc}\right] = \Sigma_\theta\left({}^Bw, {}^Ba\right), 
\end{equation}
where $\theta$ represents the learnable parameters of the covariance decoder model.

Overall, the corrected acceleration and angular velocity ${}^B{\tilde a}$ and ${}^B{\tilde w}$, are formulated as the summation of the original input ${}^Ba. {}^Bw$, the network corrections ${}^B{\sigma_{acc}}, {}^B{\sigma_{gyro}}$, and noise of the measurements:
\begin{equation}
    {}^B{\tilde a} = {}^Ba + {}^B{\sigma_{acc}} + \eta^{gyro}_k,\  {}^B{\tilde w} = {}^Bw + {}^B{\sigma_{gyro}} + \eta^{acc}_k.
\end{equation}
Combined with these outputs with the learned uncertainty $\eta^{gyro}_k, \eta^{acc}_k$, we utilize the IMU kinematic model to estimate the states ${}^Wx_j$, and more importantly, to propagate the corresponding covariance $\Sigma_{i,j}$.

\subsection{Differentiable Integration and Covariance Module}
\label{Model: Integrator}
Labeling the ground truth of the high-frequency data requires expensive motion simulators, which is infeasible due to the cost.
Therefore, AirIMU builds a differentiable integrator and covariance propagator to supervise the data-driven module throughout the integration and covariance propagation. 

The IMU pre-integration follows Newton's kinematic laws, which are well-defined in \cite{forster2015imu}. 
The integrated state $x_k$ includes orientation $R$, velocity $v$, and position $p$ within the reference frame $\{W\}$ in the $k$-th frame.
The integration from $i$-th frame to $j$-th frame is formulated as:
\begin{equation}
    \begin{split}
    \label{formula: 2.integrate}
        {\Delta}R_{ij} &= \int_{t \in [t_i, t_j]} {}^B \tilde w_k \ dt \\
        {\Delta}v_{ij} &= \int_{t \in [t_i, t_j]} R_{k} {}^B \tilde a_k \ dt \\
        {\Delta}p_{ij} &= \int\int_{t \in [t_i, t_j]} R_{k} {}^B \tilde a_k\ dt^2\ ,
    \end{split}
\end{equation}
where the ${\Delta}R_{ij}$, ${\Delta}v_{ij}$ and ${\Delta}p_{ij}$ denotes the increments of rotation, velocity and position from frame $i$ to frame $j$ respectively.
Given these increments, the $j$-th state of the robot ${}^Wx_j$ in the world frame $\{W\}$ can be predicted through:
\begin{equation}
    \begin{split}
    \label{formula: 3.predict}
        \tensor[^W]{R}{_j} &= {\Delta}R_{ij} \otimes {}^WR_i \\
        {}^Wv_j &= {}^WR_i * {\Delta}v_{ij}   + {}^Wv_i\\
        {}^Wp_j &= {}^WR_i * {\Delta}p_{ij}   + {}^Wp_i + {}^Wv_i \Delta t_{ij},
    \end{split}
\end{equation}
where the ${}^WR_i$, ${}^Wv_i$, and ${}^Wp_i$ are the state of $i$-th frame.

The propagation of the IMU covariance also follows the IMUs' kinematic model.
Following the conventions established by Forster et al. \cite{forster2015imu}, the covariance of the predicted state is characterized as:
$[\delta \phi_{k}^T, \delta v_{k}^T, \delta p_{k}^T]^T \sim \mathbb{N}(0_{9 \times 1}, \Sigma_{k})$, where the $\delta \phi_{k}$,  $\delta v_{k}$ and $ \delta p_{k}$ represents the zero-mean Gaussian noise for rotation, velocity and position.
We apply a similar covariance propagation strategy as \cite{forster2015imu} to model the covariance matrix $\Sigma_{i,i}$ of the IMU integration from frame i to frame i+1 through iterative propagation. Starting from the initial covariance matrix $\Sigma_{i,i} = 0_{9 \times 9}$, we predict the covariance matrix as:
    \begin{equation}
    \label{formula: 4.cov}
        \Sigma_{i,i+1} 
          = A \Sigma_{i,i} A^T + B_g \mathrm{diag}(\eta^{gyro}_i) B_g^T + B_a \mathrm{diag}(\eta^{acc}_i) B_a^T,
    \end{equation}
where state transition matrix A is defined as:
             \begin{equation}A =
            \begin{bmatrix}
                  {\Delta}R_{ik+1}^T & 0_{3 \times 3}  & 0_{3 \times 3} \\
                  -{\Delta}R_{ik} ({}^Ba_k^{\wedge}) {\Delta}t & I_{3 \times 3} & 0_{3 \times 3} \\
                  -1/2{\Delta}R_{ik} ({}^Ba_k^{\wedge}) {\Delta}t^2 & I_{3 \times 3} {\Delta}t & I_{3 \times 3}
                  \end{bmatrix}.
                \end{equation}
The $B$ matrix addresses the frame-by-frame uncertainty of the acceleration and the angular velocity. 
\begin{equation}
            B_g = \begin{bmatrix}
                    J_r^k \Delta t  \\
                    0_{3 \times 3}  \\
                    0_{3 \times 3} 
                \end{bmatrix}, \ \ \ 
                B_a = \begin{bmatrix}
                    0_{3 \times 3} \\
                    {\Delta}R_{ik} {\Delta}t  \\
                    1/2 {\Delta}R_{ik} {\Delta}t^2
                \end{bmatrix}.
\end{equation}
The covariance matrix is $9 \times 9$ matrix, and we denote it as:

\begin{equation}
\Sigma_{i,j+1} = 
\begin{bmatrix}
\Sigma^r_{i,j+1} & 0_{3 \times 3} & 0_{3 \times 3}\\
0_{3 \times 3} & \Sigma^v_{i,j+1} & 0_{3 \times 3} \\
0_{3 \times 3} & 0_{3 \times 3} & \Sigma^p_{i,j+1} \\
\end{bmatrix},
\end{equation}
where $\Sigma^r_{i,j+1}, \Sigma^v_{i,j+1} \text{and} \Sigma^p_{i,j+1}$ are $3\times 3$ matrix representing the covariance of rotation, velocity, and position respectively.



\subsection{Loss Function and Training strategy}
\label{Model: Loss}

To supervise the learnable model in AirIMU, we design the state loss \eqref{Eq: state loss} for noise correction and covariance loss \eqref{Eq: cov loss} for uncertainty estimation.
Given the differentiable integrator defined in \eqref{formula: 2.integrate} and \eqref{formula: 3.predict}, we train the denoising module $f_\pi(w, a)$ by supervising the state loss with the position, orientation, and translation in the world frame $\{W\}$.
\begin{equation}
\label{Eq: state loss}
    \begin{split}
        L_r &= \| \log({}^W R_i \otimes {}^W \hat R_i) \|_h, \\
        L_v &= \| {}^W v_i -  {}^W \hat v_i \|_h, \\
        L_p &= \| {}^W p_i - \hat {}^W p_i \|_h. \\
    \end{split}
\end{equation}
The $\| \cdot \|_h$ denotes the Huber function, chosen for its robustness in penalizing the large error terms, particularly in lengthy integration sequences.
We transform the orientation error to $\log$ space in the loss function \eqref{Eq: state loss} to supervise the rotation in the manifold, which reduces the single value during training.

To supervise the covariance, we use a similar loss function derived in Russell et al.  \cite{russell2021multivariate}. 
For the covariance of rotation $\Sigma^r_{i,j}$, velocity $\Sigma^v_{i,j}$ and position $\Sigma^p_{i,j}$, we design the covariance losses as:
\begin{equation}
\label{Eq: cov loss}
    \begin{split}
        L^{cov}_r &= \frac{1}{2}\left(\left\|\log({}^WR_j \otimes  {}^W \hat R_j)\right\|^2_{\Sigma^r_{i,j}}  + \ln (\det\Sigma^r_{i,j})\right), \\
        L^{cov}_v &= \frac{1}{2}\left(\left\|{}^Wv_j - {}^W \hat v_j\right\|^2_{\Sigma^v_{i,j}} + \ln (\det \Sigma^v_{i,j})\right), \\
        L^{cov}_p &= \frac{1}{2} \left(\left\|{}^W p_j - {}^W \hat p_j\right\|^2_{\Sigma^p_{i,j}}  + \ln (\det \Sigma^p_{i,j})\right), \\
    \end{split}
\end{equation}
where $y$ is the ground-truth label and $f(w, a)$ output the integrated state. 
The $\Sigma_\theta(x)$ denotes the covariance model that predicts the covariance matrix given the learned uncertainty $\eta$. 
To simplify and stabilize the training process, our training focuses on the diagonal components of the covariance matrix.
The final loss function can be denoted as:
\begin{equation}
L = L_r + L_v + Lp + \varepsilon (L^{cov}_r + L^{cov}_v +L^{cov}_p),
\end{equation}
where we set the weight of the covariance loss as $\varepsilon = 1\times 10^{-3}$.
We chose this parameter to mitigate the impact of relatively small covariance values, which otherwise lead to large covariance losses.

In contrast to previous methods that utilized data augmentation techniques such as random bias and rotation equivalence, we avoid applying data augmentation to sensor model training.
Our approach preserves the original noise in the raw data, which is critical for the AirIMU model to learn the noise pattern in the real sensor.
For most datasets, we set the length of the training segment as 1000 frames (5 seconds). 
This duration allows us to capture both the drift and the long-term features essential for IMU integration.

For the optimizer, we employ the Adam with a learning rate of $1 \times 10^{-3}$ with a weight decay $1 \times 10^{-4}$.
The weighting of each loss term and learning rate may be adjusted to accommodate the specific characteristics of different types of IMU, particularly with different bias instability of gyroscopes and accelerometers.
We will provide detailed descriptions of the parameters in the experiment sections.

\begin{figure}[!t]
	\centering
    \includegraphics[width=0.95\linewidth]{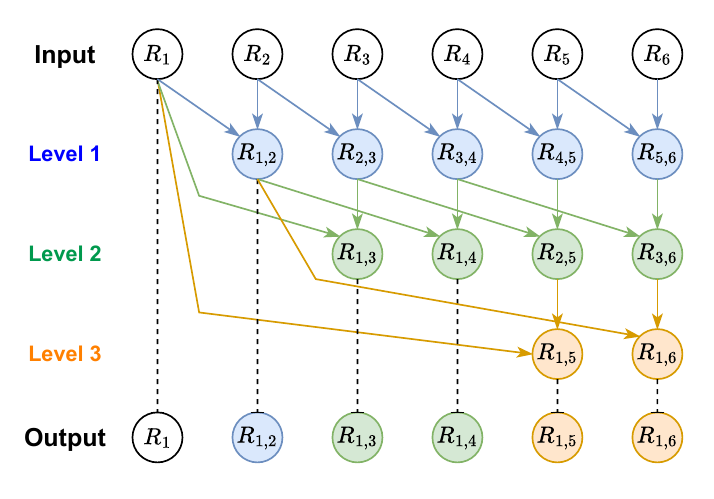}
    \caption{
    This graph presents the parallel scan we use to accelerate the cumulative product in our IMU integration. Given a sequence of gyroscope input from $R_1$ to $R_n$, the algorithm iterates through $\log(n)$ layers. In each layer $i$, it updates the last $N-2^i$ elements by multiplying the first $2^i$ elements. In this algorithm, each layer is a parallel batch that supports parallel computation.}
	\label{Fig:Tree}
\end{figure}
\begin{algorithm}[!t]
\caption{Cumulative product with $\mathcal{O}(\log N)$ complexity.}
\label{Code:Cum}
\begin{algorithmic}[1]
\State $I \gets \text{input}$; $N \gets \text{length of } I$
\For{$i \in {(0, 1, \cdots,  \log_2(N)+1)}$}
    \State $k \gets 2^{i}$
    \State $\text{indices} \gets \{k, k+1, \ldots N\}$
    \State $I[\text{indices}] \gets  I[\text{indices}] \times I[\text{indices}-k]$ 
\EndFor
\State \Return $I$
\end{algorithmic}
\end{algorithm}

\subsection{Implementation \& Acceleration}
\label{Model: Accelerate}
In our training process, we extend the supervision period to enhance the effectiveness of the training, which is particularly crucial with high-end IMUs that only exhibit drift after long periods of integration.
While extending the training over long segments can be time-consuming, our AirIMU model addresses this challenge by fully leveraging parallel computing capabilities powered by \textit{PyTorch}.
In this section, we introduce the algorithm we designed to accelerate the cumulative product and cumulative summation to facilitate parallel computation during training.
Additionally, we reformulate the iterative covariance propagation to enable batched operation for long segments. 
These efforts not only accelerate the training process but also optimize the efficiency of real-time applications.

Specifically, the IMU preintegration and covariance propagation defined in \eqref{formula: 2.integrate} and \eqref{formula: 4.cov} relies heavily on the cumulative product of $SO3$ on manifolds and cumulative summation of the acceleration.
To accelerate the integration, we design a parallel scan algorithms as shown in \fref{Fig:Tree}, where each node represents an element in $SO3$.
In this implementation, the integration with a length of $N$ can be separated into $\log_2(N)$ iterations.
During each iteration $i$, it updates each node with the preceding node positioned $2^{i}$ place before it, which can be parallel computed.
This approach achieves $O(\log N)$ time complexity while maintaining the $O(N)$ memory usage.
For more details, we show the pseudo-code of our algorithm:

Different from the segment tree strategy proposed by Brossard et al. \cite{brossard2020denoising}, which predicts only the final state $R_{1,6}$ of the gyroscope integration, our approach is designed to calculate every intermediate result throughout the integration within the same level of time.
The intermediate results significantly accelerates the calculation of the matrix of $A$, and thereby accelerates the covariance propagation.
Moreover, we not only use this approach in gyroscope integration, but also for accelormeter integration and the covariance propagation.
Overall, we integrate this algorithm into two operators: the \texttt{cumsum} for the cumulative summation and the \texttt{cumprod} for the cumulative product of the $SO3$ on manifolds.


The conventional solution outlined in \eqref{formula: 4.cov} iteratively calculates the covariance of each state.
This iterative approach, while methodically sound, requires performing covariance propagation thousands of times, even for supervising brief segments lasting a few seconds.
Such an iterative method is inefficient for training, leading to significant computational burdens. 
To address this inefficiency and fully utilize the parallel computing capabilities, we reformulate the covariance propagation for $\Sigma_{i,j}$ to two steps.

First of all, we calculate the matrix list $C^A_{i,j}$ with $\prod_{k=j}^i A_{k}$ through the \texttt{cumprod} of the matrix $A_k$,  
and the matrix list $C^B_{i,j}$ consisting the initial covariance $\Sigma_{i,i}$ and the matrix $B$.
\begin{equation}
    \begin{split}    
        C^A_{i,j} =& [\prod_{k=j}^i A_{k}, \prod_{k=j}^{i+1} A_{k},...,A_j, I_{9\times9}], \\
        C^B_{i,j} = & [\Sigma_{i,i}, B_i, ..., B_{j-1}, B_j].\\
    \end{split}
\end{equation}
Secondly, the covariance $\Sigma_{i,j+1}$ can be calculated by the product over $C^A_{i,j}$ and $C^B_{i,j}$:
\begin{equation}
    \Sigma_{i,j+1} = C^A_{i,j}  \text{diag}(C^B_{i,j})  {C^A_{i,j}}^T.
\end{equation}

With this re-implementation, we are now able to supervise the IMU integration over extended sequences. For instance, in the helicopter navigation scenario described in \sref{Exp: nearearth}, our model is trained with unusually long segments that exceed one minute in duration. This extended supervision period captures the drift inherent in the navigation-grade IMU after long-term integration, which is crucial for achieving reliable navigation in aerial environments.


\subsection{IMU-GPS PGO system}
\label{Model: System}

\begin{figure}[t]
	\centering 
    \includegraphics[width=\linewidth]{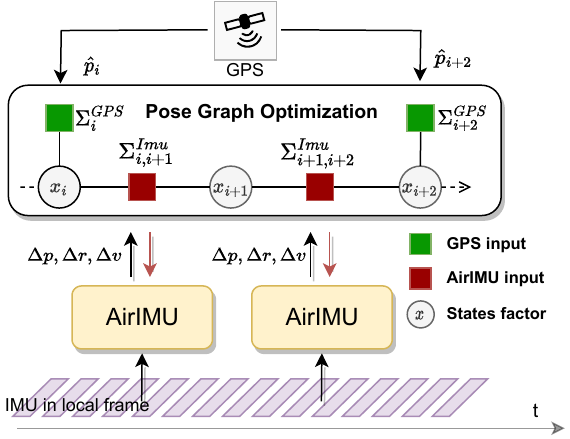}
    \caption{Pose graph optimization of a GPS signal with IMU pre-integration. The state of the graph is defined as $x_{i}$ constrained by two different observations: IMU ($\Delta p$, $\Delta v$ and $\Delta r$) and GPS ($\hat p_i$). We back-propagate the gradient through the IMU pre-integration and covariance propagation to learn the integration and noise model.}
	\label{Fig:PGOGPS}
\end{figure}
To recover the inevitable drift from inertial odometry, auxiliary sensors like GPS are employed to fuse with the IMU measurements in the PGO.
When fusing the IMU measurements with other sensor, the covariance of the IMU integration $\Sigma_\theta(x)$, plays a critical role in the fusion.
In this section, we describe the construction of a PGO that incorporates both the AirIMU model and GPS.
We demonstrate how leveraging the learned covariance from AirIMU enhances the optimization process.
This integrated system not only compensates for the drift inherent in IMU pre-integration but also effectively utilizes the complementary sensors, resulting in more reliable and accurate navigation results.

In \fref{Fig:PGOGPS}, we show the system pipeline and the factor graph incorporating GPS signals.
The objective of the PGO is to optimize the state of the robot $x$, which is modeled as the factor in our PGO.
From the AirIMU, we take the pre-integrated increments ($\Delta p$, $\Delta v$, and $\Delta r$) as the observation.
These observations are then propagated alongside the initial state $x_i (R_i, v_i, p_i)$ to calculate the state $x_j (R_j, v_j, p_j)$, forming the basis to compute the IMU residual:

\begin{equation}
\begin{aligned}
    e^{IMU}_{ij}(\Delta v_{ij}, \Delta p_{ij}, \Delta r_{ij}) = \begin{bmatrix}
        {\Delta}R_{ij} {}^WR_i  ({}^WR_j)^{-1} \\
       {}^Wv_i + {}^WR_i * {\Delta}v_{ij}  - {}^Wv_j\\
         {}^Wp_i + {}^WR_i * {\Delta}p_{ij}  - {}^Wp_j
    \end{bmatrix}.\\
    \; 
\end{aligned}
\end{equation}

To calculate the corresponding covariance matrix $ \Sigma^{IMU}_{ij} $ of the IMU integration. 
We leverage the predicted uncertainty from the network $[\eta^{acc}_k, \eta^{gyro}_k] = \Sigma_\theta(w, a)$ and employ it in \eqref{formula: 4.cov} to propagate the covariance matrix $ \Sigma^{IMU}_{ij}$.
Since the initial state $x_i$ is updated after each optimization, we also update the corresponding covariance $\Sigma_{i,i}$ of the $i^{th}$ state during the optimization.
This ensures the covariance of the state is bounded, giving the other sensors' measurements and advancing the reliability.

We define the GPS signal as $\hat p_i \in R^3$ to constrain the state of position ${}^Wp_i$, where the residual of GPS measurements are:
\begin{equation}
e^{GPS}_i( {}^W \hat p_i) = {}^W \hat p_i -  {}^Wp_i 
\end{equation}
where the measurement of the GPS is $\hat {}^Wp_i$ with a corresponding covariance matrix defined as a diagonal matrix $ \Sigma^{GPS}_i $.
The optimization with both the $e^{GPS}_i(\hat p_i)$ and $e^{IMU}_{ij}(\Delta v_{ij}, \Delta p_{ij}, \Delta r_{ij})$ can be formulated as:
\begin{equation}
X^* = \argmin_{X}\left\{ \sum_{k\in G} \|e^{GPS}_k\|^2_{\Sigma^{GPS}_k} +  \sum_{l \in L} \| e^{IMU}_l \|^2_{\Sigma^{IMU}_l} \right\}
\end{equation}
where $G$ and $L$ is the set of GPS and IMU observations and the $X$ is the factor. We can solve this optimization by Levenberg-Marquardt algorithms. 
\section{Experiments}

\begin{table}[!t]
    \caption{Datasets summary}
    \centering
    \resizebox{\linewidth}{!}{
    \begin{tabular}{l|llll}
        \toprule
        \textbf{Datasets} & \textbf{Duration} & \textbf{IMU} & \textbf{Modality}\\
        \midrule
        EuRoC \cite{burri2016euroc}   & 22m29s & ADIS16448 & Drone \\ 
        TUM-VI \cite{schubert2018tum}  & 13m31s  & BMI160 & Handheld\\
        SubtMRS \cite{zhao2023subt} & 2h52m  & Epson M-G365 & Ground robot\\
        KITTI \cite{KITTIDataset}  & 43m44s  & OXTS RT 3000 & Vehicle \\
        ALTO \cite{hemann2016long} & 2h12m  & NG LCI-1 & Helicopter\\
        \bottomrule
        \end{tabular}
        \label{datasets}
    }
\end{table}

\begin{table*}[!t]
    
    \caption{The ROE (Unit: \si{\degree}) and RPE (Unit: meter) of IMU Pre-integration over 1 second (200 frames) on the TUMVI dataset.}
    \label{Tab: TUM-Kalibr}
    \centering
    \resizebox{0.88\linewidth}{!}{
    \begin{tabular}{C{1cm}|C{1.2cm}C{1.2cm}C{1.2cm}C{1.2cm}C{1.2cm}C{1.2cm}C{1.2cm}C{1.2cm}}
        \toprule
        \multirow{2}{*}{\textbf{Seq.}} & \multicolumn{2}{c}{\textbf{Baseline}} & \multicolumn{2}{c}{\textbf{\textbf{Brossard et al.\cite{brossard2020denoising}} }}& \multicolumn{2}{c}{\textbf{Kalibr \cite{rehder2016extending} + Baseline}}& \multicolumn{2}{c}{\textbf{AirIMU} }\\
         & ROE  & RPE & ROE  & RPE& ROE  &RPE& ROE  & RPE  \\
         \midrule
         Room 2 & 2.3161 & 0.7652 & 0.7075 & - &  0.7006  & 0.0785 & \textbf{0.6765} & \textbf{0.0770}\\
         Room 4 &2.8239 &0.7558 & 0.4460 & - & 0.4397 & 0.0571 &\textbf{0.3930} &\textbf{0.0540}\\
         Room 6 & 2.3407 & 0.8521&0.4029 & - & 0.3923 &0.4096 &\textbf{0.3743}&\textbf{0.4093}\\
         \midrule
         Avg. &2.4936 &0.7910  & 0.5188& - &0.5109 & 0.1817 & \textbf{0.4813}&\textbf{0.1801} \\
        \bottomrule
    \end{tabular} }
\end{table*}

As shown in \fref{Fig:Devices}, we comprehensively evaluate our method across a full spectrum of IMUs, from low-cost automotive-grade to high-end navigation-grade IMUs, each featured on different platforms and modalities. 
The price and gyro in-run bias stability are plotted, revealing a clear correlation between cost and performance stability.
At the high end of the spectrum, we have the \texttt{NG LITEF LCI-1} in the ALTO dataset \cite{cisneros2022alto} with a cost exceeding \$50,000, showcasing exceptional bias stability in helicopters.
The tactical-grade IMUs, like the \texttt{OXTS RT3003} in the KITTI dataset \cite{KITTIDataset} and the \texttt{Epson M-G365} in the Subt-MRS \cite{zhao2023subt}, are priced at approximately \$25,000 and \$1,400, deployed on UGVs and vehicles respectively.
On the low-cost end, we feature the industrial-grade IMU \texttt{BMI 160} on drones at around \$500 and the automotive-grade IMU \texttt{ADIS 16448} at \$3 used in handheld platforms, demonstrating the feasibility of using low-cost IMUs for certain applications.
These datasets evaluate the adaptability of our model not only across multiple grades of IMUs but also across various platforms and modalities.

In our experiments, we demonstrate the dual purposes of the AirIMU: state estimation and uncertainty estimation. 

\textbf{State Estimation}
In Section \ref{Exp: Denoising}, Section \ref{Exp: Learning IO} and \ref{Exp: Subt}, we demonstrate the AirIMU model's superior performance in IMU integration, outperforming both learning-based IMU de-noising algorithms and traditional IMU calibration algorithms.
Notably, the AirIMU model achieves a tenfold increase in accuracy compared to learning-based inertial odometry methods in short-term integration.
In Section \ref{Exp: nearearth}, we demonstrate the potential of the AirIMU model in GPS-denied helicopter navigation using high-end IMUs. 
This assessment spans a cruising distance over 262 $\si{\km}$, providing a realistic and challenging environment for evaluating our method.

\textbf{Learned Uncertainty:}
In  Section \ref{Exp: PGO} and Section \ref{Exp: Subt}, we assess our model with a particular emphasis on the potency of the learned uncertainty.
An ablation study in Section \ref{Exp: PGO}, focused on a GPS-PGO experiment, reveals an average improvement of 31.6\% in utilizing the learned uncertainty.
Furthermore, in Section \ref{Exp: Subt}, we conduct an ablation study to highlight the benefits of jointly training the uncertainty module with the IMU noise correction.
This dual training approach significantly improves the network's ability to extract features from IMUs, resulting in a performance enhancement of 17.8\% compared to models solely trained on noise correction.

To establish a baseline for comparison, we implement the IMU pre-integration using PyPose \cite{wang2023pypose} based on the work of Foster et al. \cite{forster2015imu}, which we denoted as \textbf{Baseline}. 
This serves as a reference for evaluating the performance of IMU integration with raw data.
In our experiments, each dataset contains IMUs with diverse mechanical conditions and modalities. Since each sensor features a distinct noise model, training a universal model that can generalize across all IMUs and platforms without compromising accuracy is infeasible. 
Therefore, we train distinct sensor models tailored to different IMUs.

\subsection{Evaluation Metrics}
We define the following metrics to evaluate the performance of our model. It is important to note that the relative error is calculated across an assigned time interval like 1s.

\textbf{Relative Orientation Error} (ROE, $\si{\radian}$) is defined as
\begin{equation}
    \text{ROE} = \frac{1}{N} \sum^N_{i=1}\left\| \log\left( \hat{R}_{i,i+\Delta t}^T  R_{i,i+\Delta t}\right)\right\|_2, 
\end{equation}
where $R_{i,i+\Delta t}$ is a rotation over the assigned time intervals $\Delta t$,  calculated as $R_{i,i+\Delta t}= R_{i}^T R_{i+\Delta t} $. ROE calculates the average of all fixed-duration orientation errors.

\textbf{Relative Position Error} (RPE, $\si{\meter}$) is defined as
\begin{equation}
    \text{RPE} = \frac{1}{N}\sum^N_{i=1}{\left\| p_{i+\Delta t} - p_{i} - R_{i}\hat{R}_{i}^T \left(\hat{p}_{i+\Delta t} - \hat{p_{i}}\right)\right\|_2},
\end{equation}
calculates the average errors in relative displacements over the assigned time intervals. The ground truth poses from the world frame are aligned with the integrated poses.

\textbf{Rotational Root Mean Squared Error} (R-RMSE, $\si{\radian}$),
\begin{equation}
     \text{R-RMSE} = \sqrt{\frac{1}{N} \sum^N_{i=1}\left\| \log\left(\hat{R}_{i,i+\Delta t}^T R_{i,i+\Delta t}\right)\right\|^2_2},
\end{equation}
calculates the RMSE of the rotations over the assigned time intervals on the manifold space.

\textbf{Positional Root Mean Squared Error} (P-RMSE, $\si{\meter}$)
\begin{equation}
    \text{P-RMSE} =\sqrt{ \frac{1}{N}\sum^N_{i=1}\left\| p_{i+\Delta t} - p_{i} - R_{i}\hat{R}_{i}^T \left(\hat{p}_{i+\Delta t} - \hat{p_{i}}\right)\right\|^2_2},
\end{equation}
calculates the RMSE of the relative displacements over the assigned time intervals.


\textbf{Absolute Translation Error} (ATE, $\si{\meter}$)
\begin{equation}
    \text{ATE} = \frac{1}{N}\sum^N_{i=1}{\left\| p_{i} - \hat{p_{i}}\right\|_2},
\end{equation}
calculates the mean of absolute distance between the estimated positions and the ground truth over all time points.


\subsection{IMU Pre-integration Accuracy}\label{Exp: Denoising}
In this section, we evaluate the accuracy of the AirIMU model in IMU pre-integration.
We compare against two existing learning-based IMU denoising methods: Zhang et al. \cite{zhang2021imu} and Brossard et al. \cite{brossard2020denoising} on the EuRoC dataset \cite{burri2016euroc}. 
Additionally, we showcase how our method enhances the integration precision of the traditional calibration algorithm Kalibr \cite{rehder2016extending} when tested on the TUM-VI dataset \cite{schubert2018tum}.

\textbf{TUM-VI Dataset}:
As described in \ref{sec: imu-denoise}, traditional methods typically employ a linear model to calibrate IMU intrinsic, such as bias. In the TUM-VI dataset \cite{schubert2018tum}, IMU was calibrated using Kalibr \cite{rehder2016extending} as a reference.
We leverage the calibrated data to examine whether our AirIMU model can further improve traditional IMU calibration methods.

To benchmark the TUM-VI dataset, we select \texttt{room 1}, \texttt{room 3}, and \texttt{room 5} for training, while using the remaining sequences for evaluation. 
It's important to note that this dataset is not entirely collected within the Vicon system.
Therefore, we mask the segments that are not observable by the Vicon during training.
To ensure fairness, we train Brossard et al.'s model with the same training set and show the results on the same evaluation set.
We evaluate the ROE and RPE over 1-second intervals to assess gyroscope and accelerometer integration. 
Gauging relative orientation and translation offers a more consistent metric for short-term integration, given the nature of noise accumulation over time in IMU integration. 


In the experiment, Kalibr effectively corrects sensor offsets using linear models.
Grounded on these results, AirIMU improves upon this by modeling the non-deterministic error in a data-driven manner, demonstrating a further improvement compared to the Kalibr. 
As shown in \tref{Tab: TUM-Kalibr}, AirIMU exhibits robustness and superiority in improving IMU integration accuracy. 
For the orientation, the average ROE of AirIMU is 0.4813$^{\circ}$, which is superior to 2.4936$^{\circ}$ from the raw-data integration, 0.5188$^{\circ}$ from Brossard et al. and 0.5109$^{\circ}$ from Kalibr. 
For relative translation, AirIMU achieves an average RPE of 0.1801\si{\meter}, better than the 0.7910\si{\meter} from raw-data integration and the 0.1817\si{\meter} from Kalibr.

\textbf{EuRoC Dataset}: 
In this section, we evaluate the pre-integration performance of AirIMU with existing learning-based IMU de-noising methods \cite{zhang2021imu, brossard2020denoising}.
These studies were assessed using different visual-inertial odometry and evaluation benchmarks.
Specifically, Zhang et al. passed the preprocessed IMU measurements to their own visual-inertial algorithm for evaluation.
On the other hand, Brossard et al. passed their corrections to the Open-VINS framework \cite{geneva2020openvins} for visual-inertial fusion.
As a result, It is infeasible to directly compare absolute positional error, which makes it difficult to assess the effectiveness of the IMU denoising module.

Considering these differences, in our evaluation, we compare each method separately using the benchmarks proposed in their papers, respectively.
Zhang et al. reported the results of P-RMSE and R-RMSE over ten frames of IMU measurements. 
Thus, we evaluate 10-frame IMU pre-integration results in \tref{Tab: IMUpre}.
For Brossard et al., we show both the R-RMSE and ROE in \tref{Tab: IMUgyro}.
To ensure fairness, we split the training and testing set in the same way as Zhang et al. and Brossard et al.

\begin{table}[H]
    \caption{IMU integration result of 10 frames on EuRoC Dataset with P-RMSE (Unit: \si{\cm}) and R-RMSE (Unit: \si{\degree})}
    \label{Tab: IMUpre}
    \centering
    \resizebox{\linewidth}{!}{
    \large
    \begin{tabular}{C{1cm}|C{1.19cm}C{1.19cm}C{1.19cm}C{1.19cm}C{1.19cm}C{1.19cm}}
        \toprule
        \multirow{2}{*}{\textbf{Seq.}} & \multicolumn{2}{c}{\textbf{Zhang et al.} \cite{zhang2021imu}}  & \multicolumn{2}{c}{\textbf{Baseline}} & \multicolumn{2}{c}{\textbf{AirIMU}} \\
        
         & {\small R-RMSE} & {\small P-RMSE}  & {\small R-RMSE} & {\small P-RMSE}  & {\small R-RMSE} & {\small P-RMSE}\\
         \midrule
         MH02 &  0.143 & 0.280 & 0.230 & 0.046 & \textbf{0.020} & \textbf{0.040}\\
         MH04 & 0.212 & 0.580 & 0.229 & 0.326 & \textbf{0.025} & \textbf{0.325} \\
         V103 & 0.384 & 0.530 & 0.231 & 0.048 & \textbf{0.036} & \textbf{0.042} \\
         V202 & 0.390 & 0.830 & 0.240 & 0.028 & \textbf{0.037} & \textbf{0.027} \\
         V101 & 0.928 & 0.300 & 0.228 & 0.030 & \textbf{0.021} & \textbf{0.020} \\
         \midrule
         Avg.  & 0.411 & 0.504 & 0.232 & 0.096 & \textbf{0.028} & \textbf{0.091} \\
        \bottomrule
    \end{tabular} 
    }
\end{table}

\begin{figure}[H]
	\centering
    \includegraphics[width=\linewidth]{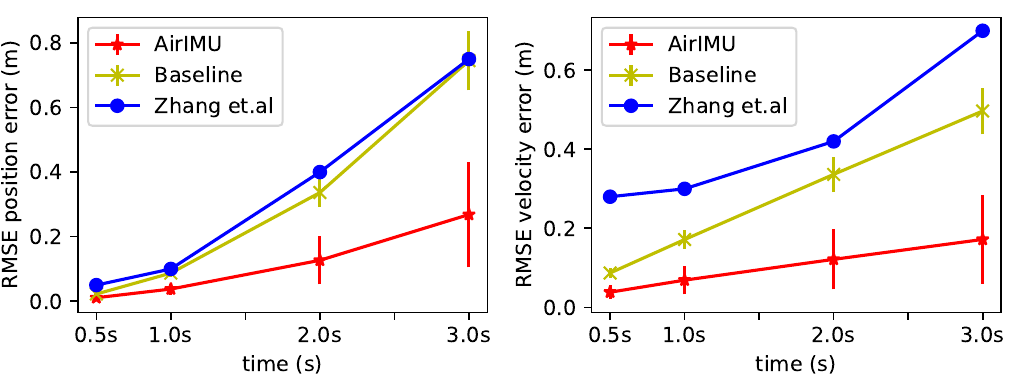}
    \caption{
        We present the RMSE of both translation and velocity over intervals of 0.5\si{s}, 1\si{s}, 2\si{s}, and 3\si{s}. The results illustrate the accumulation of errors throughout the integration, where AirIMU remains accurate after integration.
        }
	\label{Fig: IMUpreInterval}
\end{figure}

Notably, our method achieves a ten-fold improvement in accuracy compared to their reported P-RMSE and R-RMSE for short-term integration.
Moreover, Zhang et al. provided P-RMSE values over specific time intervals: 0.5, 1, 2, and 3 seconds. Their reported values were approximately 0.05, 0.10, 0.40, and 0.75 meters, respectively. In contrast, using the same intervals, our method achieved P-RMSE values of 0.011, 0.038, 0.127, and 0.27 meters. This translates to an average improvement of three times in terms of positional accuracy. Additionally, we present the velocity RMSE results in \fref{Fig: IMUpreInterval}.

Since Brossard et al. reported their findings using ROE while Zhang et al. reported R-RMSE, we show both the R-RMSE and ROE in our evaluation for consistency with prior experiments. Results over the one-second intervals (200 frames) are shown in \tref{Tab: IMUgyro}.

\begin{table}[H]
    \caption{Gyroscope integration on EuRoC Dataset, we show the R-RMSE and the ROE (Unit: \si{\degree}).}
    \centering
    \label{Tab: IMUgyro}
    \resizebox{\linewidth}{!}{
    \begin{tabular}{C{0.7cm}|C{0.7cm}C{0.6cm}C{0.9cm}C{0.9cm}C{0.6cm}C{0.6cm}}
        \toprule
        \multirow{2}{*}{\textbf{Seq.}}  & \multicolumn{2}{c}{\textbf{Baseline}} & \multicolumn{2}{c}{\textbf{Brossard et al.} \cite{brossard2020denoising}}  & \multicolumn{2}{c}{\textbf{AirIMU}} \\
         &RMSE & ROE & RMSE & ROE & RMSE & ROE\\
         \midrule
         MH02  & 4.5800 & 4.5799 &  0.1255 & 0.0871 & \textbf{0.0973} & \textbf{0.0789} \\
         MH04 & 4.5406 & 4.5391 &  0.3556 & 0.1067  & \textbf{0.0836} & \textbf{0.0708} \\
         V103 & 4.4909 & 4.4870 & 0.2181 & 0.1935  & \textbf{0.2107} & \textbf{0.1884} \\
         V202 & 4.7000 & 4.6924  & 0.2595 & 0.2389 & \textbf{0.2366} & \textbf{0.2157} \\
         V101  & 4.5275 & 4.5252 &  \textbf{0.1346} & \textbf{0.1173} & 0.1413 & 0.1241 \\
         \midrule
         Avg.  &4.5678  & 4.5647 & 0.2189 & 0.1487 & \textbf{0.1305} & \textbf{0.1127}\\
        \bottomrule
    \end{tabular}
    }
\end{table}

In the \tref{Tab: IMUgyro}, AirIMU shows superior accuracy and reliability on IMU gyroscope integration. 
The Baseline's average R-RMSE and ROE are 4.5678$^{\circ}$ and 4.5647$^{\circ}$, respectively.
These errors are primarily due to a significant gyroscope offset in the EuRoC dataset, which causes notable drift. In contrast, AirIMU demonstrates significantly better performance with R-RMSE and ROE values of 0.1305$^{\circ}$ and 0.1127$^{\circ}$, respectively, representing a ten-fold improvement over the Baseline. Moreover, when compared to Brossard et al., AirIMU consistently achieves better results across most sequences, with improvements of 40.3\% in R-RMSE and 24.2\% in ROE on average.

\subsection{Learning-based inertial odometry}\label{Exp: Learning IO}

In \sref{Sec: LIO}, we identified two main approaches to data-driven IO. 
The first utilizes end-to-end trained networks to estimate the velocity.
Within this category, we examine the estimation results derived from the ResNet, which is prevalently employed in methods such as RoNIN \cite{herath2020ronin}.
The second approach involves EKF-fused inertial odometry like TLIO \cite{liu2020tlio}.
When integrating the neural network's predictions with IMU signals, the EKF in TLIO adeptly addresses system biases, enhancing accuracy.

In this section, we evaluate the integration result of the AirIMU model and the estimation results from the learning inertial odometry. 
In the pedestrian dataset, the subject is highly correlated with the motion patterns.
As the walking pace typically shows periodic acceleration cycles and the velocity of the pedestrian is constant in most scenarios, models like RoNIN and TLIO can explicitly estimate the subject's velocity by motion patterns.
However, this assumption does not hold in autonomous driving datasets like KITTI and UAV navigation datasets like EuRoC, since both vehicles and drones exhibit different velocities and multiple moving patterns.
Therefore, we compare our model against existing learning inertial odometry algorithms, including RoNIN and TLIO, on the EuRoC and KITTI datasets.

\textbf{EuRoC Dataset}:
We assessed the P-RMSE over an interval of 200 frames (equivalent to 1s), comparing these values to the short-term integration outcomes from AirIMU and Baseline in \tref{Tab: LIO}.
Compared to the output from the directly regressed neural network, the pre-integration results from AirIMU demonstrate a substantially greater accuracy — by a factor of 17. 
This distinction is especially noticeable when we consider the drone data. 
Unlike the pedestrian dataset that possesses clear motion patterns or modalities, the dynamics of the drone remain unpredictable, making it challenging for the network to accurately predict the velocity.

\begin{table}[t]
    
    \caption{The P-RMSE of IMU integration over 1 second (200 frames) on EuRoC dataset (Unit: meter).}
    \label{Tab: LIO}
    \centering
    \resizebox{\linewidth}{!}{
    \begin{tabular}{C{0.7cm}|C{1.3cm}C{1.5cm}C{1.3cm}C{1.3cm}}
        \toprule
        {\textbf{Seq.}} & {\textbf{Baseline}} & {\textbf{RoNIN} \cite{herath2020ronin}} & {\textbf{TLIO}\cite{liu2020tlio}} & {\textbf{AirIMU}} \\
        
         \midrule
         MH02 & 0.1899 & 0.5186 & 0.3030 & \textbf{0.0234} \\
         MH04 & 0.1904 & 1.0377 & 0.8230 & \textbf{0.0415} \\
         V103 & 0.2116 & 0.7387 & 0.6530 & \textbf{0.0583} \\
         V202 & 0.1520 & 0.7497 & 0.5090 & \textbf{0.0851} \\
         V101 & 0.1903 & 0.4964 & 0.3430 & \textbf{0.0363} \\
         \midrule
         Avg. & 0.1868 & 0.7082 & 0.5262 & \textbf{0.0489} \\

        \bottomrule
    \end{tabular} 
    }
\end{table}






\textbf{KITTI Dataset}:
To demonstrate the generalizability across different collection dates, we employ our algorithms on the KITTI dataset\cite{KITTIDataset}, renowned for its state estimation benchmarks with vision, IMU, and LiDAR.
We train the AirIMU on the KITTI Odometry datasets, specifically collected on \texttt{2011\_09\_30} and \texttt{2011\_10\_03}. 
To evaluate the data from different collection dates, we conduct the quantitative analysis and qualitative comparisons on the KITTI raw datasets collected from \texttt{2011\_09\_26} and \texttt{2011\_09\_28}.
This setup ensures the AirIMU's robustness across temporal variations.

To adjust the low IMU frequency characteristic of the KITTI dataset, the GRU layer in the AirIMU model is removed. Furthermore, we set the window size for pre-integration to 5 seconds during the training, introducing supervision on long-term integration. Notably, the trajectories 0016 (from \texttt{2011\_09\_28}) and 0057 (from \texttt{2011\_09\_26}) are removed from the test dataset due to the misaligned ground truth poses verified by manual inspection.

The ROE and RPE for Baseline, RoNIN, and AirIMU on the test sequences are shown in \tref{Tab: KITTI-raw}. Quantitative results show that the AirIMU outperforms the Baseline method on most trajectories and demonstrates a more than 20 times lower RPE than the RoNIN. 
Remarkably, the RoNIN model shows an exceptionally high error compared to the Baseline method. This is because the vehicle's speed changes rapidly, which makes it difficult for the network to predict the velocity. 

\begin{table}[t]
    \caption{The ROE (Unit: \si{\degree}) and RPE (Unit: meter) of IMU integration over 1 second (10 frames) on the test dataset. RoNIN refers specifically to the ResNet-50 model\cite{herath2020ronin}.}
    \label{Tab: KITTI-raw}
    \centering
    \resizebox{\linewidth}{!}{
    \begin{tabular}{C{0.7cm}|C{.8cm}C{.8cm}C{.7cm}C{.9cm}C{.8cm}C{.8cm}}
        \toprule
        \multirow{2}{*}{\textbf{Seq.}} & \multicolumn{2}{c}{\textbf{Baseline}} & \multicolumn{2}{c}{\textbf{RoNIN}} & \multicolumn{2}{c}{\textbf{AirIMU}}\\
         & ROE  & RPE & ROE  & RPE & ROE & RPE \\
        \midrule
        0014 & 0.4137 & 0.1785 & - & 12.2424 & \textbf{0.3927} & \textbf{0.1699}\\
        0018 & 0.2451 & \textbf{0.0715} & - & 3.8301 & \textbf{0.2388} & 0.0857\\
        0022 & 0.5319 & 0.2893 & - & 4.9544 & \textbf{0.4912} & \textbf{0.2841}\\
        0029 & 0.3435 & 0.393  & - & 6.6598 & \textbf{0.3245} & \textbf{0.3562}\\
        0035 & 0.3682 & 0.0969 & - & 2.0726 & \textbf{0.3563} & \textbf{0.0956}\\
        0039 & \textbf{0.5273} & 0.1703 & - & 4.4768 & 0.5514 & \textbf{0.1613}\\
        0106 & 0.2052 & 0.1568 & - & 8.8109 & \textbf{0.2026} & \textbf{0.1512}\\
        \midrule
        Avg. & 0.3764 & 0.1938 & - & 4.8909 & \textbf{0.3653} & \textbf{0.1863}\\
        \bottomrule
    \end{tabular} }
\end{table}

\begin{figure}[t]
    \centering
    \includegraphics[width=\linewidth]{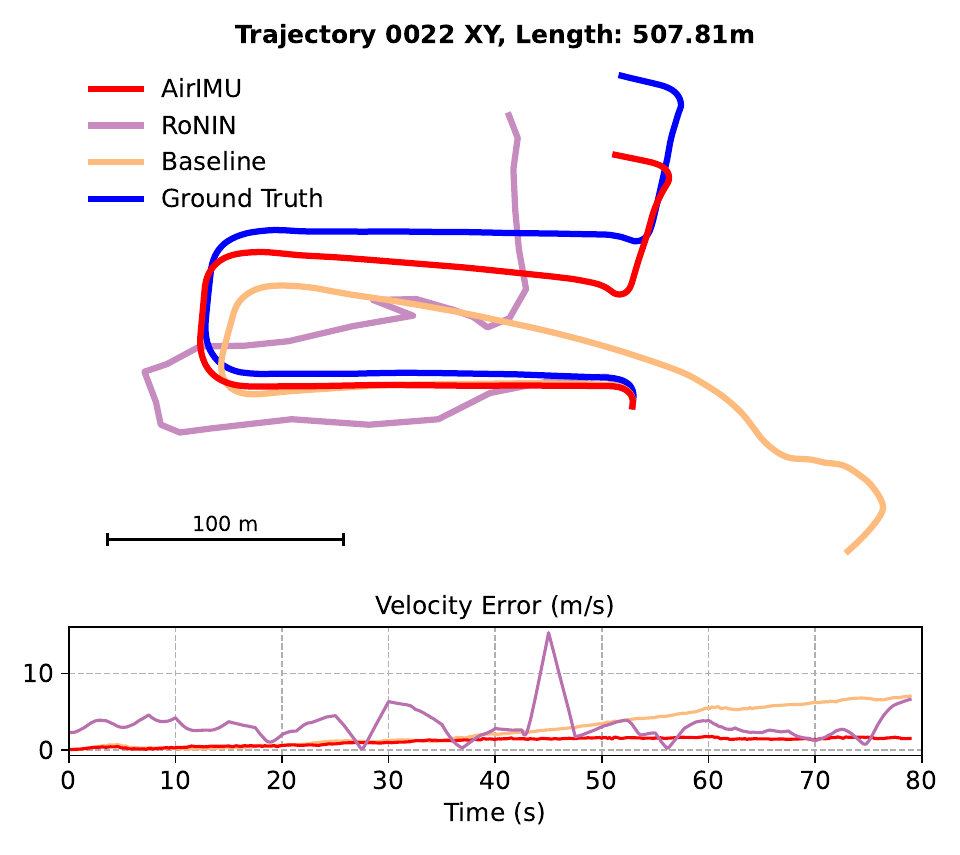}
    \caption{
        Trajectory 0022 in KITTI-Raw dataset estimated by {\color{roninpink}RoNIN (purple)}, {\color{orange}Baseline (orange)} and {\color{red}AirIMU (red)}. 
        The graph below shows the accumulated velocity error (Unit: \si{\meter\per s}) through the entire trajectory. AirIMU significantly alleviates the velocity error accumulation on the y-axis compared to the Baseline.
        As a result, AirIMU maintains its accuracy after integration, while the Baseline method exhibits noticeable drift during the third turn.
        }
	\label{Fig: FigKITTI}
\end{figure}


\begin{table*}[t]
    \caption{Ablation study of the GPS PGO with 1 Hz on the EuRoC dataset with different covariance settings.}
        \label{Tab: Ablation1Hz}
    \centering
    \resizebox{\linewidth}{!}{
    \begin{tabular}{C{3.6cm}|C{2cm}C{2cm}C{2cm}C{2cm}C{2cm}}
        \toprule
        {\textbf{Seq.}} & {{MH02}} & {{MH04}} & {{V103}} & {{V202}}  & {{V101}} \\
         \midrule
        {Raw IMU w/ VinsMono Cov}\raggedright 
        & 0.1493 $\pm$ 0.0044 & 0.1491 $\pm$ 0.0050 & 0.1549 $\pm$ 0.0058          & 0.1542 $\pm$ 0.0051          & 0.1478  $\pm$ 0.0041 \\
        {AirIMU w/ VinsMono Cov}\raggedright 
        & 0.1484 $\pm$ 0.0044          & 0.1486 $\pm$ 0.0049         & 0.1547 $\pm$ 0.0057          & 0.1540  $\pm$ 0.0049    &0.1469 $\pm$ 0.0039\\
        {Raw IMU w/ learned Cov}\raggedright 
        &  0.1184 $\pm$ 0.0041         & 0.1327 $\pm$ 0.0048         & 0.1310 $\pm$ 0.0051          &  0.1569 $\pm$ 0.0078     & 0.1265 $\pm$ 0.0055\\
        {AirIMU w/ learned Cov}\raggedright 
        & \textbf{0.1026 $\pm$ 0.0035} & \textbf{0.1024 $\pm$ 0.0039} & \textbf{0.1039 $\pm$ 0.0028} & \textbf{0.1060 $\pm$ 0.0046} & \textbf{0.1019 $\pm$ 0.0041}\\

        \bottomrule
 \multicolumn{4}{c}{Results show the mean and the standard deviation with ATE in meter \si{\meter}.}\\
    \end{tabular} 
    }
\end{table*}

\begin{table*}[t]
    \caption{Ablation study of the GPS PGO with 0.1 Hz on the EuRoC dataset with different covariance settings.}
        \label{Tab: Ablation01Hz}
    \centering
    \resizebox{\linewidth}{!}{
    \begin{tabular}{C{3.6cm}|C{2cm}C{2cm}C{2cm}C{2cm}C{2cm}}
        \toprule
        {\textbf{Seq.}} & {{MH02}} & {{MH04}} & {{V103}} & {{V202}}  & {{V101}} \\
         \midrule
        {Raw IMU w/ VinsMono Cov}\raggedright & 2.1454 $\pm$ 0.0150         & 3.6112 $\pm$ 0.0255         & 1.4474 $\pm$ 0.0309          & 1.6840 $\pm$ 0.0165          & 1.4740  $\pm$ 0.0178 \\
        {AirIMU w/ VinsMono Cov}\raggedright & 1.9235 $\pm$ 0.0181          & 0.2955 $\pm$ 0.0270         & 0.3085 $\pm$ 0.0289          & \textbf{0.8681  $\pm$ 0.0214 }   &0.3455 $\pm$ 0.0161\\
        {Raw IMU w/ learned Cov}\raggedright &  2.5088 $\pm$ 0.0368         & 4.3450 $\pm$ 0.0280         & 1.4248 $\pm$ 0.0227          &  2.1213 $\pm$ 0.0184     & 1.1941 $\pm$ 0.0141\\
        {AirIMU w/ learned Cov}\raggedright & \textbf{0.2414 $\pm$ 0.0240} & \textbf{0.2754 $\pm$ 0.0274} & \textbf{0.2905 $\pm$ 0.0216} & 0.8927 $\pm$ 0.0189 & \textbf{0.2948 $\pm$ 0.0144}\\

        \bottomrule
 \multicolumn{4}{c}{Results show the mean and the standard deviation with ATE in meter \si{\meter}.}\\
    \end{tabular} 
    }
\end{table*}

\begin{figure*}[t]
	\centering
    \includegraphics[width=\linewidth]{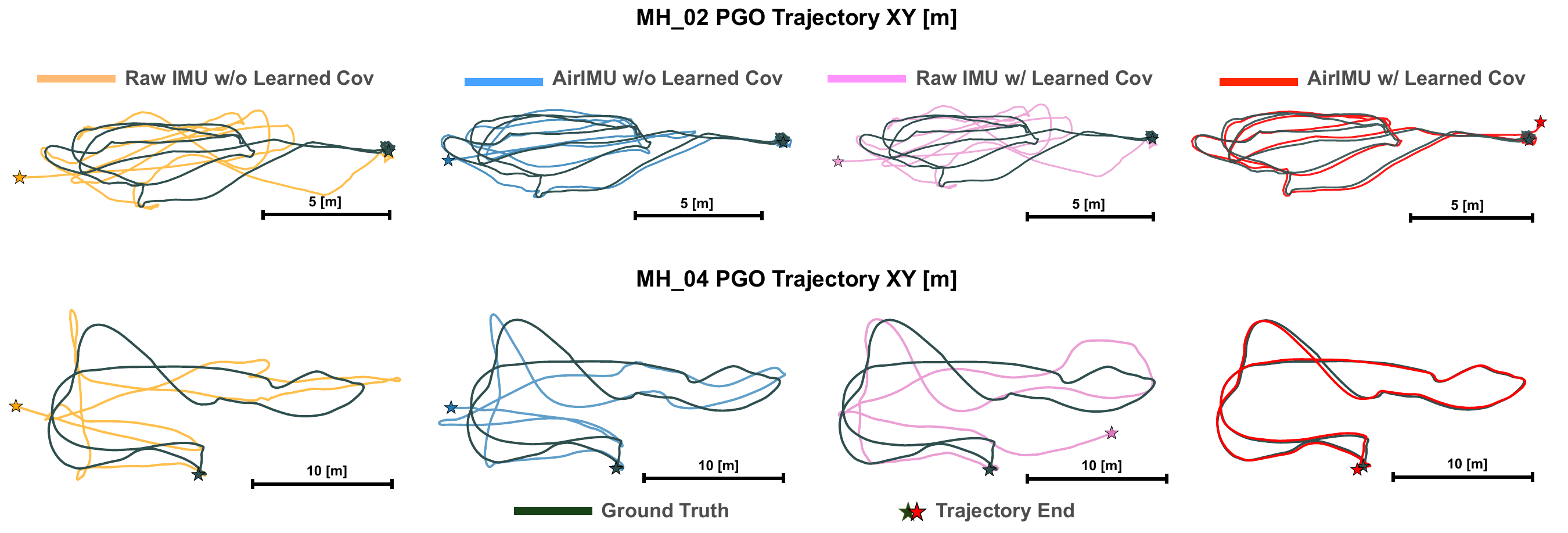}
    \caption{Qualitative analysis of IMU GPS PGO results. In this experiment, the GPS frequency is set to 0.1 Hz. Despite using identical IMU measurements, AirIMU with a learned covariance yields optimal fusion results.} 
	\label{Fig: PGOVisualize 1}
\end{figure*}


The trajectory 0022 from \texttt{2011\_09\_26} estimated by Baseline, RoNIN, and AirIMU, is plotted in \fref{Fig: FigKITTI} along with the velocity error over the entire trajectory for qualitative analysis. 
The trajectory estimated by AirIMU significantly reduces the velocity error compared to the Baseline method. On the other hand, the Baseline tends to quickly accumulate errors in both velocity and orientation when the vehicle is turning, while AirIMU alleviates this problem.
We also highlight the deficiency of learning-based IO in the purple trajectory. Due to the lack of prior knowledge of initial velocity, the RoNIN model shows multiple unexpected sharp turns on straight-motion segments, while AirIMU maintains its accuracy.

Combining the data-driven module and kinematic model, AirIMU outperforms the learning IO over a long sequence in KITTI.
In the demonstrated trajectory, the ATE estimated by AirIMU (30.40 \si{\meter}) is only half of that estimated by the Baseline method (60.14 \si{\meter}), and the RoNIN model (48.19 \si{\meter}).
Overall, these results demonstrate both the accuracy and the temporal generalizability of the AirIMU.

\subsection{IMU-GPS PGO}\label{Exp: PGO}
The uncertainty model of the IMU pre-integration is pivotal when fusing the IMU measurements with other sensors, such as the GPS.
As shown in \tref{Tab: IMUpre}, our AirIMU model can achieve high accuracy in the short-time pre-integration. 
The integrated trajectory, however, will drift in long-term integration due to the unavoidable noise inherent in the sensor. 
Therefore, when fusing the IMU pre-integration with low-frequency auxiliary measurements, the propagated covariance model determines how well the optimization can trust the IMU measurements.
We conduct an ablation study with different IMU sensor models and covariance setups to assess the effectiveness of our learned uncertainty proposed in the AirIMU model.
In the ablation study, we utilize the GPS-PGO system we formulated in \fref{Fig:PGOGPS}, where GPS was considered as an unary constraint to the factor.

A comprehensive evaluation of the IMU covariance requires controlling the noise generated from the fused sensors.
To this end, we generate simulated GPS signals with Gaussian noise of $\sigma_{GPS} = 0.1$ \si{\meter} at a frequency of 1 Hz in \tref{Tab: Ablation1Hz}. 
This setup aims to replicate a typical RTK-GPS signal and allows us to evaluate the baseline performance of the AirIMU covariance under standard operating scenarios.
Furthermore, to explore the model's efficacy in less ideal conditions—such as unstable GPS signals or situations with limited observational data (e.g., loop closure scenarios), we also simulate GPS signals at a lower frequency of 0.1 Hz in \tref{Tab: Ablation01Hz}. 
This setup is designed to test the effectiveness of our learned uncertainty model when faced with sporadic or diminished GPS data inputs, thereby demonstrating its adaptability and robustness in challenging navigational environments.

\begin{figure*}[t]
	\centering
    \includegraphics[width=\linewidth]{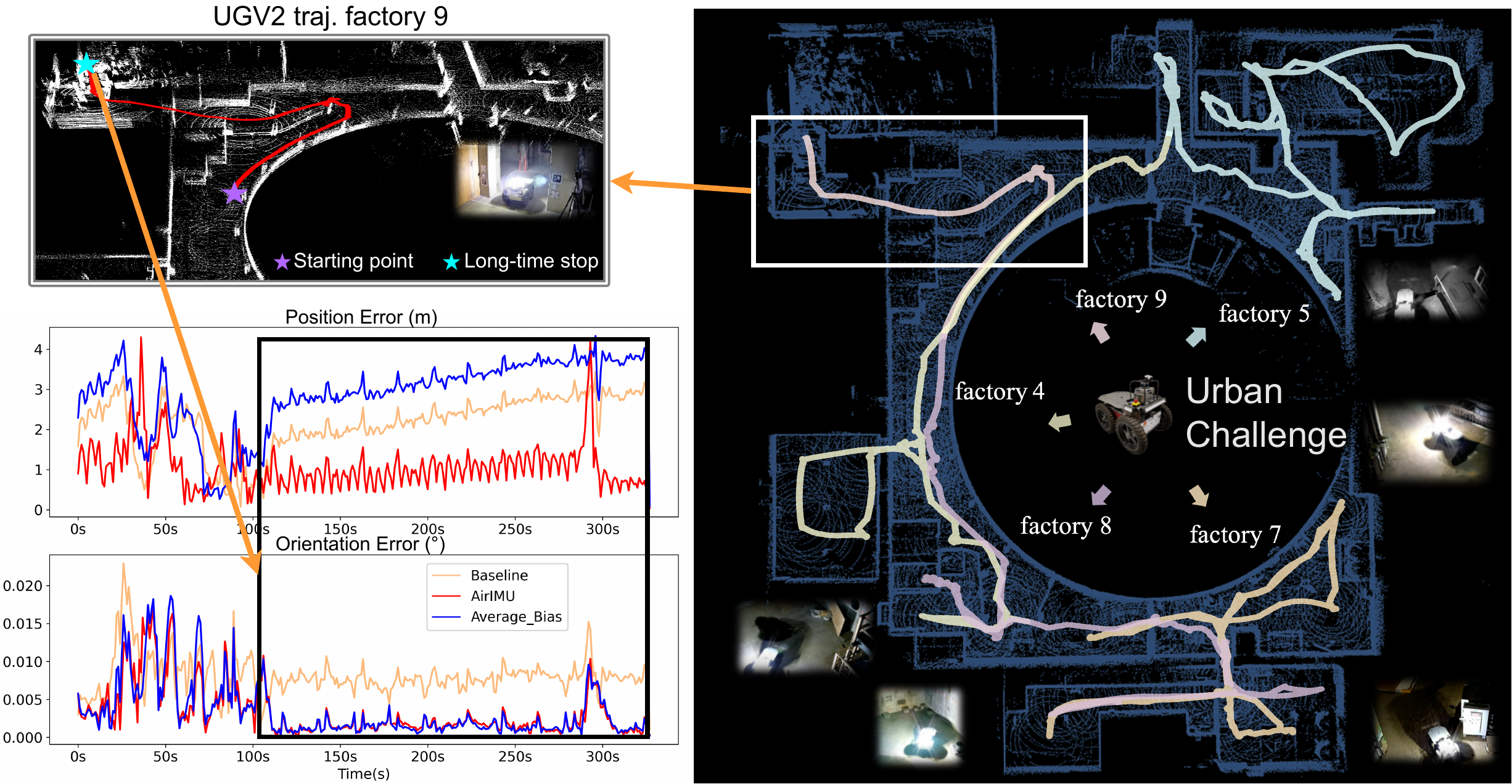}
    \caption{We select the sequence \texttt{factory 9} from UGV2 in the DARPA SubT Urban Circuit experiments. This sequence captures the robot navigating over an obstacle and remains stationary for a period. During the stationary period (highlighted with white box), \textbf{\iref{subt:b} Average Bias} degrades the integration performance compared to raw-data integration. This is because the bias patterns differ when the robot is stationary. \textbf{AirIMU} successfully captures the biases while the IMU is stationary.}
	\label{Fig:SubtQualitative}
\end{figure*}

The PGO experiment with the following settings:

\begin{enumerate}[(a)]
\item Raw IMU with covariance from VinsMono's setting \cite{qin2018vins}. \label{PGO: a}
\item AirIMU with the covariance from the VinsMono's setting.\label{PGO: b}
\item Raw IMU with proposed learning covariance model.\label{PGO: c}
\item AirIMU with proposed learning covariance model.\label{PGO: d}
\end{enumerate}
The proposed covariance model of \textbf{{\iref{PGO: d}}} is propagated through the Equation. \ref{formula: 4.cov} with the learned uncertainty model in $\Sigma_\theta(w,a)$. 
The covariance matrix of \textbf{{\iref{PGO: b}}} is calculated by the VinsMono's EuRoc configuration, where the noise standard deviation of accelerometer $\eta^{\text{acc}} = 0.08$ and the noise standard deviation of gyroscope $\eta^{\text{gyro}} = 0.004$.
These parameters are later applied in the equation \ref{formula: 4.cov} to calculate the covariance matrix of the IMU pre-integration $\Sigma^{IMU}$.
The covariance matrix of the GPS signals is defined to be the inverse of a diagonal matrix: 

$$\Sigma_k^{GPS} = \mathrm{diag}([1/\sigma^2_{GPS}, 1/\sigma^2_{GPS}, 1/\sigma^2_{GPS}])$$
To ensure a fair comparison, we repeated the experiment 10 times with randomly generated GPS signals.
The average ATE and standard deviation are summarized in \tref{Tab: Ablation01Hz} and \tref{Tab: Ablation1Hz}.
Compared to \textbf{{\iref{PGO: b}} AirIMU with VinsMono}  setups, \textbf{{\iref{PGO: d}} AirIMU model with its learned covariance} significantly enhances the ATE of the optimization results with 21.6\%. 
Compared to \textbf{\iref{PGO: a} raw IMU with VinsMono's covariance}, \textbf{\iref{PGO: d} AirIMU with the learned covariance} reduces the ATE from 0.1505 to 0.1033, improving 31.6$\%$.

In \tref{Tab: Ablation01Hz}, we explore the impact of unstable GPS update conditions, specifically 1 GPS reading every 10 seconds. 
Due to this low GPS frequency (0.1 Hz), IMU must be integrated over long intervals, leading to an accumulation of integration errors. 
This presents significant challenges for sensor fusion, affecting overall system accuracy.
\textbf{\iref{PGO: d} AirIMU with learned covariance} reduces the error from 2.0724 to 0.3990, improving 80.7\% in ATE.

In \fref{Fig: PGOVisualize 1}, we plot the optimized trajectory of \texttt{MH04} and \texttt{MH02}, where the drone flies through the machine hall and returns to its original position. 
The yellow trajectory \textbf{\iref{PGO: a}} and the blue trajectory  \textbf{\iref{PGO: b}}, which employ referenced covariance parameters, have a large drift in the experiment. In these two trajectories, the orientation of the trajectory endpoint is the opposite compared to the ground truth.
In contrast, the pink trajectory \textbf{\iref{PGO: c}}, which employed learned covariance with raw IMU data, shows a marked improvement in navigational accuracy and a smaller return drift.
Overall, our model \textbf{\iref{PGO: d}} with the learned covariance illustrates the best optimization performance, maintaining high accuracy and successfully returning to the starting position.
These results demonstrate the decisive role of our learned covariance in the PGO.

\subsection{Ablation Study on Multi-robot Underground Dataset}\label{Exp: Subt}

In this section, we demonstrate the synergistic effect achieved through the joint training of the covariance model and the correction model.
We evaluate our model on the SubT-MRS dataset\cite{zhao2023subt}, a multi-robot, multi-modal, and multi-degraded challenging dataset designed to push SLAM towards all-weather environments.

Our experiments selected the multi-agent UGV datasets from the LiDAR-inertial track, collected in the DARPA Subterranean (SubT) Challenge's Final Event and Urban Circuit Event. These agents' platform is equipped with Epson M-G365 IMU. 
In the Final Challenge, three agents (UGV1, UGV2, UGV3) explored three uncharted underground environments: tunnel, urban, and cave spaces. 
In the Urban Challenge, two agents (UGV1 and UGV2) explored an unfinished nuclear power plant. 
We divided the data from different agents into sequences based on the operational scenarios, such as corridors, warehouses, and long-time stops. The duration of each sequence ranges approximately from 4 to 10 minutes.

\begin{table*}[!t]
    \caption{The RPE (Unit: meter) of IMU Pre-integration over 5 seconds (1000 frames) on SubT-MRS.}
    \label{Tab:SubT-MRS-RPE}
    \centering
    \resizebox{0.85\linewidth}{!}{
    \begin{tabular}{C{1.2cm}|C{0.6cm}C{2cm}C{1.5cm}C{1.5cm}C{1.5cm}C{1.5cm}}
        \toprule
        \multirow{2}{*}{\textbf{Datasets}} & \multirow{2}{*}{\textbf{Robots}} & \multirow{2}{*}{\textbf{Sequences}} &\multirow{2}{*}{\textbf{Baseline}} & \multirow{2}{*}{\textbf{\textbf{Average Bias} }}& \multirow{2}{*}{\parbox{1.3cm}{\centering\textbf{AirIMU}\\ \textbf{(w/o Cov)}}}& \multirow{2}{*}{\parbox{1.1cm}{\centering\textbf{AirIMU}}}\\
        &&&&&&\\
         \hline \noalign{\smallskip}
         {\multirow{7}{*}{\parbox{1.1cm}{\centering\textbf{Final} \\ \textbf{Challenge}} }}  & {UGV1}  &warehouse 1 & 2.7934 &\textbf{1.8828} & 2.2540 & 2.2306\\[1.5pt]
         \cline{2-7} \noalign{\smallskip}
           & \multirow{3}{*}{UGV2}  &tunnel room & 1.0965&1.0578 & \textbf{0.7447} &0.7568 \\
           &   &tunnel stop 4 &1.0169 &1.1519 & \textbf{1.0280} & 1.0709\\
           &   &tunnel corridor 3& 1.4248 &1.2095& \textbf{0.8479}& 0.8745\\[1.5pt]
         \cline{2-7} \noalign{\smallskip}
          & {\multirow{2}{*}{UGV3}}  &cave corridor 3& 3.5067 &1.8609 & 1.8290 &\textbf{1.6627} \\
          & &tunnel corridor 5& 2.7762 &1.3399 & 1.4243 &\textbf{1.2495}\\[1.5pt]
        \hline\noalign{\smallskip}
        \multirow{5}{*}{\parbox{1.3cm}{\centering\textbf{Urban} \\ \textbf{Challenge}}}  & UGV1 & factory 2
&4.1769&2.9747&2.8173&\textbf{2.6525}\\[1.5pt]
        \cline{2-7}\noalign{\smallskip}
         & \multirow{3}{*}{UGV2} & factory 4&2.8622&2.3964&2.2895&\textbf{1.3393}\\
          & &factory 6&1.6454&2.0282&1.5293&\textbf{0.9805}\\
          & &factory 9 &2.1193&2.8264&1.7837&\textbf{0.7534}\\[1.5pt]
        \hline\noalign{\smallskip}
         &\multicolumn{2}{c}{\textbf{Average}} &2.3418&1.8729&1.6517& \textbf{1.3571}\\[1.5pt] 
        \bottomrule
    \end{tabular} 
        }
\end{table*}

\begin{table*}[!t]
    \caption{The ROE (Unit: \si{\degree}) of IMU Pre-integration over 5 seconds (1000 frames) on SubT-MRS.}
    \label{Tab:SubT-MRS-ROE}
    \centering
    \resizebox{0.85\linewidth}{!}{
    \begin{tabular}{C{1.2cm}|C{0.6cm}C{2cm}C{1.5cm}C{1.5cm}C{1.5cm}C{1.5cm}}
        \toprule
        \multirow{2}{*}{\textbf{Datasets}} & \multirow{2}{*}{\textbf{Robots}} & \multirow{2}{*}{\textbf{Sequences}} &\multirow{2}{*}{\textbf{Baseline}} & \multirow{2}{*}{\textbf{\textbf{Average Bias} }}& \multirow{2}{*}{\parbox{1.3cm}{\centering\textbf{AirIMU}\\ \textbf{(w/o Cov)}}}& \multirow{2}{*}{\parbox{1.1cm}{\centering\textbf{AirIMU}}}\\
        &&&&&&\\
         \hline \noalign{\smallskip}
         {\multirow{7}{*}{\parbox{1.1cm}{\centering\textbf{Final} \\ \textbf{Challenge}} }}  & {UGV1}  &warehouse 1 & 2.5365&0.3604 & 0.4908 & \textbf{0.3558}\\[1.5pt]
         \cline{2-7} \noalign{\smallskip}
           & \multirow{3}{*}{UGV2}  &tunnel room &0.3887&0.1875 & 0.1796 &\textbf{0.1731} \\
           &   &tunnel stop 4 & 0.3681 &0.0853 & 0.0877 & \textbf{0.0692}\\
           &   &tunnel corridor 3 & 0.4028 &0.2125&0.1982 & \textbf{0.1907}\\[1.5pt]
         \cline{2-7} \noalign{\smallskip}
          & {\multirow{2}{*}{UGV3}}  &cave corridor 3&2.1469 &\textbf{0.3104}& 0.4129 &0.3330\\
          & &tunnel corridor 5 &2.1634&0.3597 & 0.4806 &\textbf{0.3588}\\[1.5pt]
        \hline\noalign{\smallskip}
        \multirow{5}{*}{\parbox{1.3cm}{\centering\textbf{Urban} \\ \textbf{Challenge}}}  & UGV1 & factory 2
&0.6389&\textbf{0.3302}& 0.3425&0.4613\\[1.5pt]
        \cline{2-7}\noalign{\smallskip}
         & \multirow{3}{*}{UGV2} & factory 4&0.9198&0.7144&0.7188&\textbf{0.5285}\\
          & &factory 6&0.4585&\textbf{0.1648}&0.1735&0.2157\\
          & &factory 9 &0.4618&0.1969&0.2018&\textbf{0.1865}\\[1.5pt]
        \hline\noalign{\smallskip}
        & \multicolumn{2}{c}{\textbf{Average}} &1.0485 &0.2922 &0.3286 & \textbf{0.2873}\\[1.5pt] 
        \bottomrule
    \end{tabular} 
        }
\end{table*}

To quantify the effectiveness of the covariance-aware module in the IMU noise corrections, we conduct experiments with several different setups:
\begin{enumerate}[(i)]
\item \textbf{Baseline:} \label{subt:a} This setup serves as our control, integrating the raw IMU data.
It helps to benchmark the other setups.
\item \textbf{Average Bias:} \label{subt:b} We correct the IMU data by calibrating the biases in the accelerometer and gyroscope from the training dataset. This strategy is widely adopted in traditional IMU calibration algorithms, setting a foundation for comparing with learning-based methods.
\item \textbf{AirIMU w/o Cov:} \label{subt:c} In this configuration, the AirIMU network is solely trained to correct IMU noise without supervising the covariance explicitly. This setup tests the network's basic capabilities in noise correction.
\item \textbf{AirIMU with Cov:} \label{subt:d} This advanced setup explicitly supervises the uncertainty in the IMU integration. 
It demonstrates the synergistic effect of jointly training the covariance module with the IMU noise correction.

\end{enumerate}

During the training stage, we randomly segment the sequences and group them with the same robots in the Urban Challenges and the Final Challenges into training sets and testing sets.
Since each robot in this dataset had a different calibration process before the challenge, we cannot fit a general IMU model for every sequence. 
Therefore, we train different models for different robots in each challenge.
Unlike the previous experiment, we don't prioritize the rotation error because the gyroscope of the Epson IMU has relatively good in-bias stability.
So, we pick an equal weight ratio of $1:1:1$ for position, velocity, and rotation losses.
In the Final Challenge, we initiate the learning rate at $1\times 10^{-4}$.
However, for the Urban Challenge, we increase the learning rate to $1\times 10^{-3}$ due to the smaller size of the training data set and the more complex bias distribution.

\begin{figure*}[!t]
	\centering
    \includegraphics[width=\linewidth]{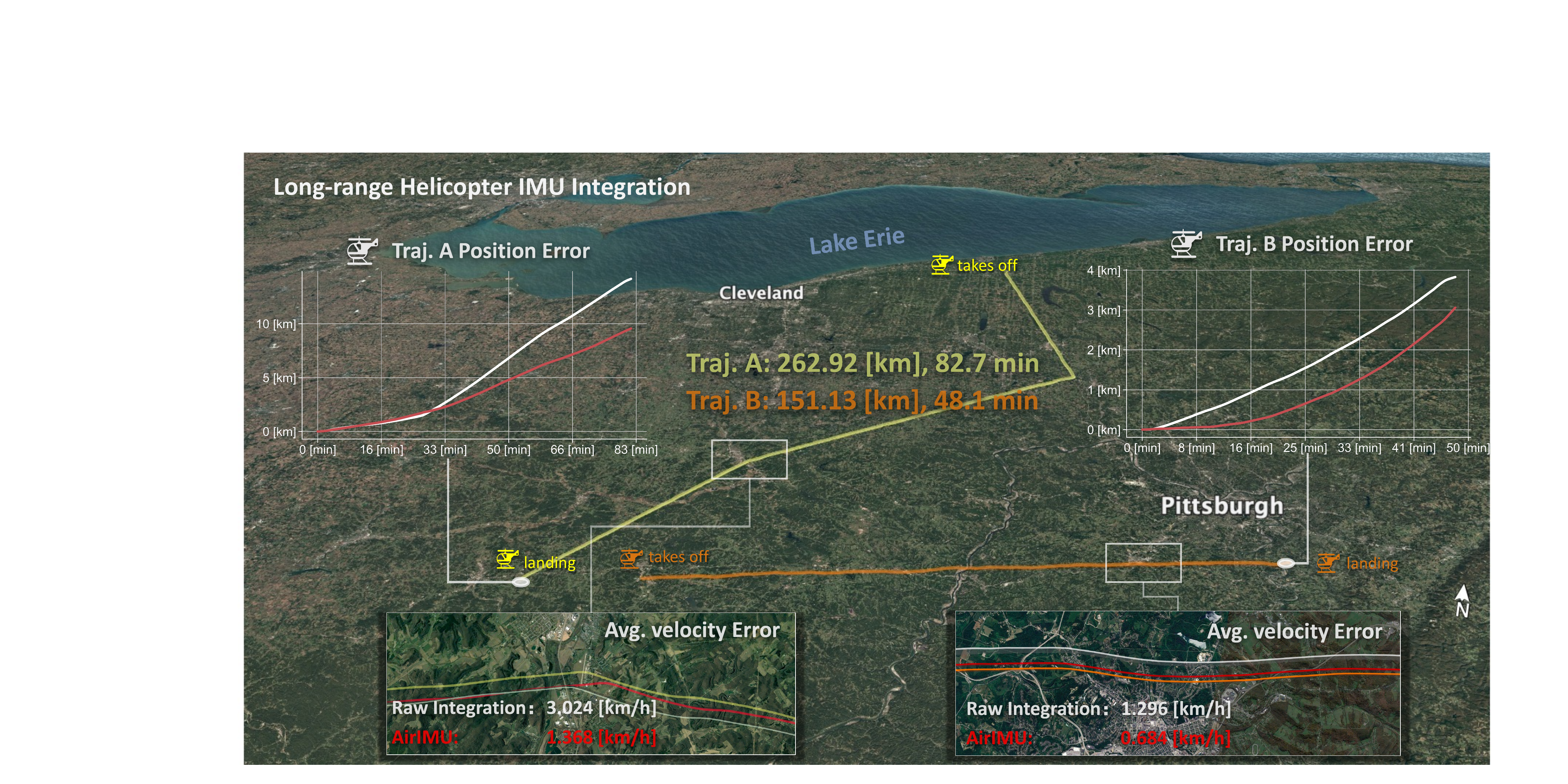}
    \caption{Trajectories from the ALTO Dataset \cite{cisneros2022alto}: raw-IMU is represented by a white line and AirIMU by a red line. 
    In these trajectories, AirIMU clearly reduces position errors.
    We also compare the average velocity error, where AirIMU significantly reduces the error for 42.8\% in Traj. A and the 50\% in Traj. B.
    }
	\label{Fig: NearEarth}
\end{figure*}

To see how well our method can compete with the linear model, we set up \textbf{\iref{subt:b} Average Bias} as a reference.
In this setting, we use \texttt{Adam} optimizer to learn the average bias of the gyroscope and the accelerometer in the training sequences. 
Later, we apply the average bias to the raw IMU of the testing sequences to integrate the states.
This strategy is similar to the classical IMU calibration algorithm \cite{rehder2016extending}, where the bias is calculated from the calibration trajectories with batched optimization.
To evaluate the benefits of jointly training the covariance module with the IMU corrections, we design the \textbf{\iref{subt:c} AirIMU w/o Cov}. In this setup, we remove the covariance supervision and solely train the AirIMU model with the state loss. This ablation study can validate the synergic effect of jointly training the covariance module. with the correction module. Overall, these control setups help to evaluate the effectiveness of AirIMU under variance conditions.

In  \tref{Tab:SubT-MRS-RPE} and \tref{Tab:SubT-MRS-ROE}, jointly training AirIMU with covariance propagation \textbf{\iref{subt:d} AirIMU W/ Cov} demonstrates the best result among the experiment groups.
This improvement may be attributed to the fact that supervising the covariance alongside the correction module improves the network's ability in feature extraction, and thereby benefits each other.
Compared to \textbf{\iref{subt:a} Baseline}, our proposed \textbf{\iref{subt:d} AirIMU w/ Cov} has an average improvement of 72.6\% in ROE and 42.1\% in RPE. 
It outperforms the \textbf{\iref{subt:b} Average Bias} for 1.69\% in ROE and 27.5\% in RPE. Remarkably, compared to \textbf{\iref{subt:c} AirIMU W/O Cov}, it improves 12.6\% in the ROE and 18.2\% in the RPE on average.
Although our model has significant improvements, the performance of the UGV1 in the Final Challenge does not have the same level of improvement as the other robots.
This is mainly because the training sequences of UGV1 are relatively short, which fails to provide enough supervision.

In \fref{Fig:SubtQualitative}, we qualitatively analyze the sequence \texttt{factory 9} and show the RPE and ROE across the entire trajectory.
In this sequence, the vehicle starts in the factory and ends in the storage room for 3 minutes, where it remains stationary, waiting for further instruction.
During the stationary period, the IMU shows a different bias pattern compared with the other sequences that are in motion. 
In the long-time stop period, the \textbf{\iref{subt:b} Average Bias} degrades the performance compared to the \textbf{\iref{subt:a} Baseline} with the raw-IMU integration.
This is because the sensors' bias in the training set doesn't have the same distribution as the testing sequences.
During the calibration stage, the bias model memorizes the average offset of the training set. But when the robot stops, the IMU has a different bias distribution.
Using \textbf{\iref{subt:b} Average Bias} has 50\% more error in the position.
While our proposed model \textbf{\iref{subt:a} Baseline} significantly reduces the positional error.
This experiment demonstrates the stability and accuracy of AirIMU, especially when the testing sequences have different distributions from the training sets.
Moreover, the ablation study shows the synergistic effect of covariance supervision and noise correction in AirIMU.

\subsection{Large-scale Helicopter IMU pre-integration}
\label{Exp: nearearth}

Many existing learning-based IOs are evaluated indoors or on small-scale datasets with limited duration and speeds. 
To bridge this gap, we employed Visual Terrain Relative Navigation dataset ALTO\cite{cisneros2022alto} from a Bell 206L (Long-Ranger) helicopter to evaluate the performance of the AirIMU model in high-end navigation grade IMU. 
As shown in \fref{Fig: NearEarth}, this dataset encompasses flights over 414 km, featuring average speeds over 150 \si{\km\per\hour} and cruising heights between 200 and 500 meters. This presents a robust testing ground for the AirIMU model's capabilities.

During the model training, we select Flight B as the training set and Flight A as the testing set, where each flight in this dataset has a duration exceeding 40 minutes. 
Given the long cruising time, we seek to expedite training and enhance robustness by removing the GRU layer from our feature encoder. 
It's worth noting that only minimal drift is observed, owing to the high precision of the navigation-grade IMU sensor (\texttt{NG LCI-1}).
We utilize the AirIMU to integrate the entire trajectory, assessing integration quality via speed error. 
Landing drift is also measured by comparing the integrated trajectory's endpoint with the actual landing spot. 
As shown in \fref{Fig: NearEarth}, compared to the Baseline method, the AirIMU model demonstrates remarkable enhancements. For Flight A, there's a speed error reduction of 42.8\% (from 3.02 \si{\km\per\hour} to 1.37 \si{\km\per\hour}). Flight B experiences a decrease of 50.0\% (from 1.29 \si{\km\per\hour} to 0.68 \si{\km\per\hour}), proving the AirIMU model's efficiency in estimating helicopter velocities.
Regarding the landing drift shown in \fref{fig:eyecatcher}, the AirIMU model significantly curtails discrepancies. For Flight A, the drift reduces from 3968 \si{\meter} to just 1064 \si{\meter}. Flight B sees a reduction from 1228 \si{\meter} to 479 \si{\meter}. This experiment showcases the adaptability of the AirIMU in large-scale inertial navigation tasks.



\subsection{Time Analysis}\label{Exp: Time}
An efficient differentiable integrator with covariance propagation is critical to training and deploying AirIMU on large-scale datasets. 
In this section, we evaluate the time consumption of the integrator we proposed in \sref{Model: Accelerate}. 
We set up the experiments into four groups:
\begin{enumerate}[(a)]
\item AirIMU integrator with batched covariance integrator
\item AirIMU integrator without covariance integrator
\item Iterative IMU integrator with covariance propagation
\item Iterative IMU integrator without covariance propagation
\end{enumerate}

For each group, we calculate the time consumption of the integration with different lengths of IMU inputs spanning from 1 to 1000 frames. 
We repeat the experiment 200 times and calculate the mean and standard deviation of time consumption.
We do the experiment on the platform with Intel i9 CPU.

\begin{figure}[t]
	\centering
    \includegraphics[width=\linewidth]{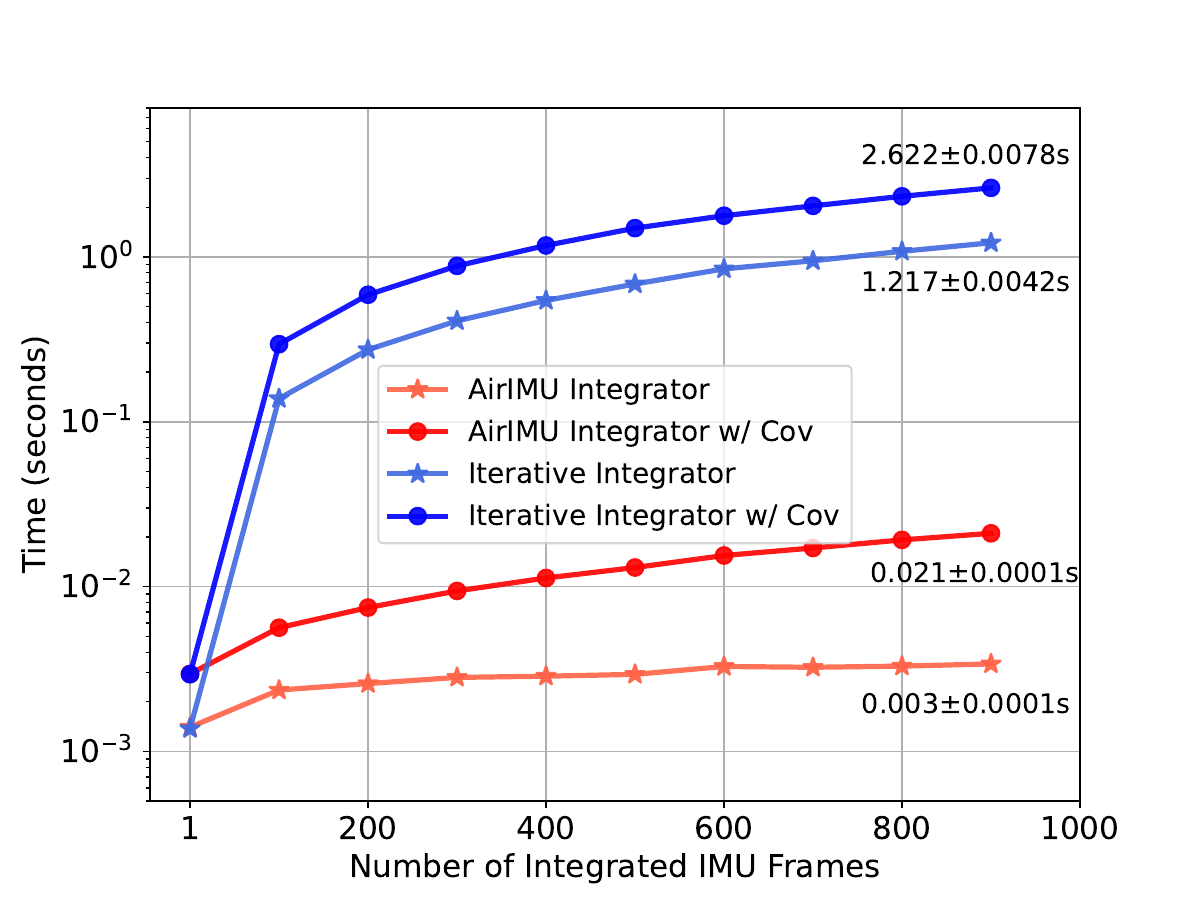}
    \caption{Time analysis of the AirIMU integrator with the integrator in an iterative manner. We integrate IMU input with different lengths (from 1 frame to 1000 frames) and display the time consumption in the $\log$ scale. The AirIMU integrator demonstrates a hundredfold improvement in efficiency.}

	\label{Fig:TimeAnalysis}
\end{figure}

As shown in \fref{Fig:TimeAnalysis}, the average time for the iterative integrator to integrate 1000 frames is 1.217s, and the average time for the AirIMU integrator is 0.021s, which is 60 times more efficient. Moreover, we compare the time consumption for the integrator with covariance propagation, where iterative covariance propagation takes 2.622s, and AirIMU with covariance propagation only takes 0.021s.
Overall, the AirIMU integrator demonstrates a hundredfold improvement in time consumption, which improves the efficiency of training and deploying AirIMU for inertial navigation.

\section{Limitation and Discussion}

In this paper, we propose a hybrid method that can adapt to a full spectrum of IMUs. However, we've observed that the AirIMU model trained on a single IMU does not readily generalize to a different IMU with different specifications and operating frequencies.
For instance, a model trained for the tactical-grade IMU \texttt{Epson G365}, used in the SubT-MRS dataset, does not apply to the automotive-grade IMU \texttt{ADIS16448} in the TUMVI dataset due to their different physical design.
Our current model is specialized, focusing on a single IMU's unique noise profile and uncertainty characteristics rather than attempting to create a one-size-fits-all model.
If we need to deploy the AirIMU model to a new type of IMU, fine-tuning with new IMU data is necessary.

In the future, we may explore the potential for generalizing through different sensors within the same models or types of IMUs.
For example, whether the sensor model trained on one \texttt{Epson G365} can be adapted to other \texttt{Epson G365} IMUs, potentially a more versatile application to scale up in manufacturing the same types of IMU.

\section{Conclusion}
We propose AirIMU, a hybrid IO system with a differentiable integrator and covariance propagator. 
To demonstrate the efficacy of the learned covariance model, we conduct an ablation study on a GPS PGO experiment, revealing that the learned covariance decreased the loss of PGO by 31\%. 
We evaluate our model in all spectrums of IMUs' grades, spanning the entry-level IMUs (Automotive Grade) to the high-end IMUs (Navigation Grade). 
Our experiments demonstrate that the AirIMU model achieves outstanding performance across the board. The evaluation also includes IMUs operating in a wide range of environments and featuring multiple agent modalities. 
We find that AirIMU not only excels at regular indoor environments but also performs well with a ground vehicle dataset over 2 \si{km} and a large-scale helicopter dataset over 414 \si{km}.
Additionally, we open-source the differentiable IMU integrator and the covariance propagation, which is implemented in the \textit{PyPose} framework.
Given these advancements, we are optimistic that the proposed method can provide a strong foundation for further research in inertial navigation systems and multi-sensor fusion.


{
\small
\bibliographystyle{./bibliography/IEEEtran}
\bibliography{./bibliography/papers}
}

{
\normalfont
\vspace{-1cm}

\begin{IEEEbiography}
[{\includegraphics[width=1in,height=1.25in,clip,keepaspectratio]{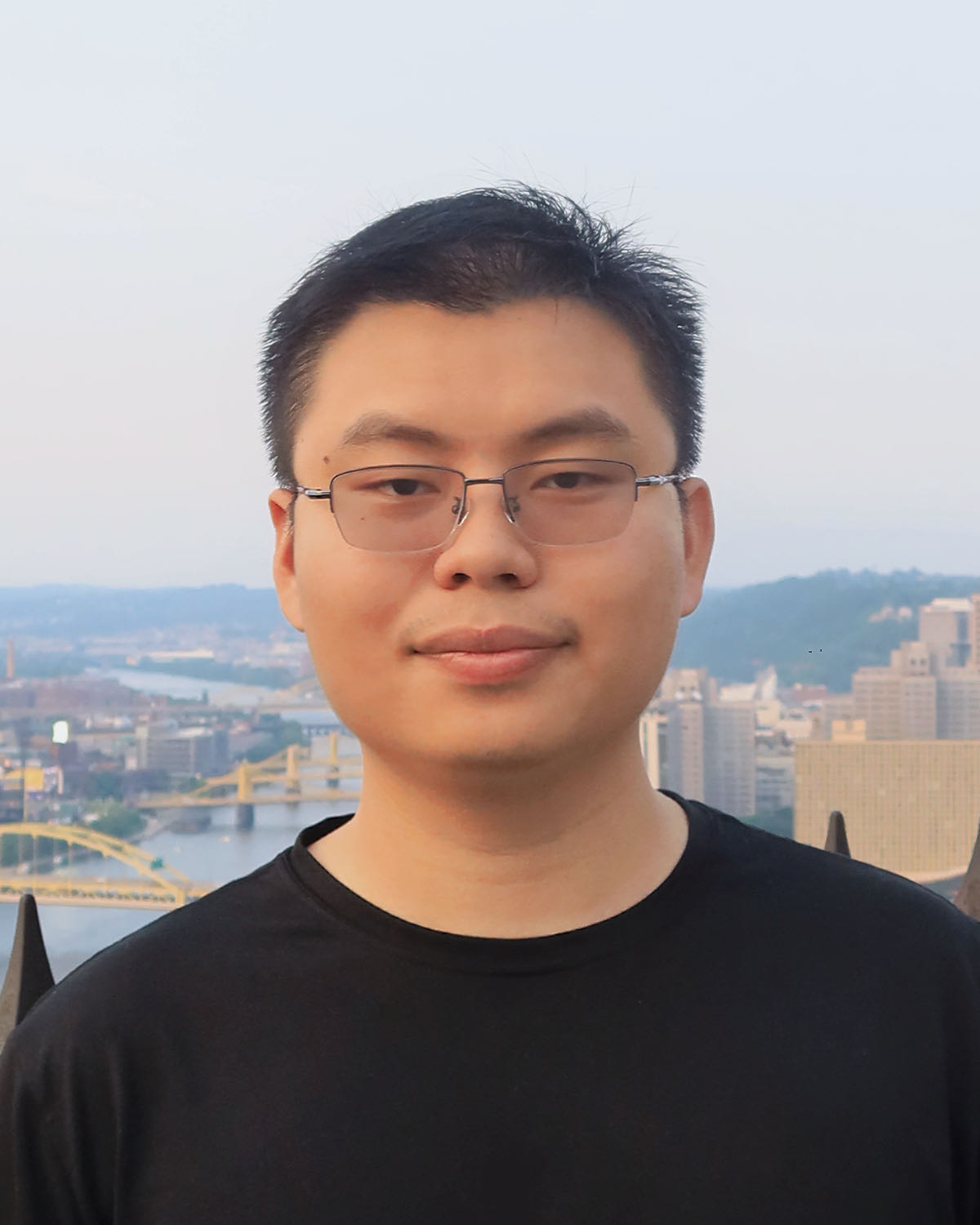}}]
{Yuheng Qiu} received the B.Eng. degree in Computer Science and Engineering from The Chinese University of Hong Kong, Shenzhen in 2019. 

He is currently a Ph.D. candidate with the Department of Mechanical Engineering at Carnegie Mellon University. His research interests include robot learning, SLAM, and inertial odometry.

He was a Reviewer for several IEEE conferences and journals, such as IEEE Robotics and Automation Letters, IEEE International Conference on Robotics and Automation, IEEE/RSJ International
Conference on Intelligent Robots and Systems, IEEE/CVF Computer Vision and Pattern Recognition Conference, and International Conference on Computer Vision.

\end{IEEEbiography}

\begin{IEEEbiography}
[{\includegraphics[width=1in,height=1.25in,clip,keepaspectratio]{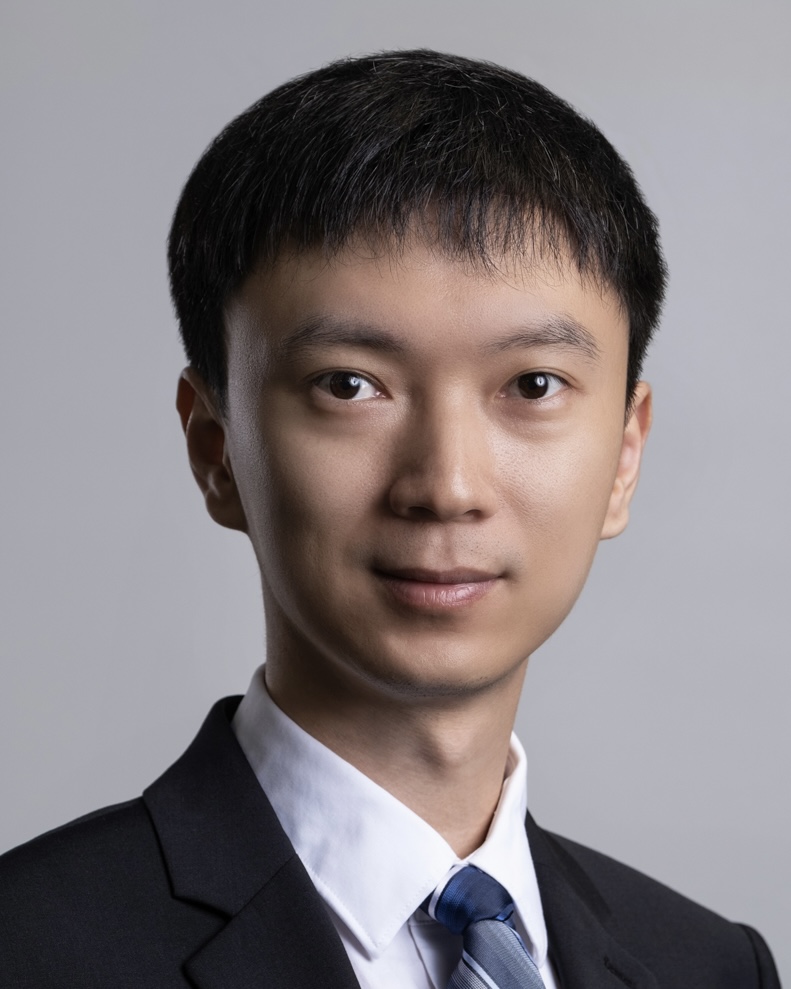}}]
{Chen Wang} received a B.Eng. degree in Electrical Engineering from Beijing Institute of Technology (BIT) in 2014 and a Ph.D. degree in Electrical Engineering from Nanyang Technological University (NTU) Singapore in 2019. He was a Postdoctoral Fellow with the Robotics Institute at Carnegie Mellon University (CMU) from 2019 to 2022.

Dr. Wang is an Assistant Professor and leading the Spatial AI \& Robotics (SAIR) Lab at the Department of Computer Science and Engineering, University at Buffalo (UB). He is an Associate Editor for the International Journal of Robotics Research (IJRR) and IEEE Robotics and Automation Letters (RA-L) and an Associate Co-chair for the IEEE Technical Committee for Computer \& Robot Vision. He served as an Area Chair for the IEEE/CVF Conference on Computer Vision and Pattern Recognition (CVPR) 2023, 2024. His research interests include Spatial AI and Robotics.
\end{IEEEbiography}

\begin{IEEEbiography}[{\includegraphics[width=1in,height=1.25in,clip,keepaspectratio]{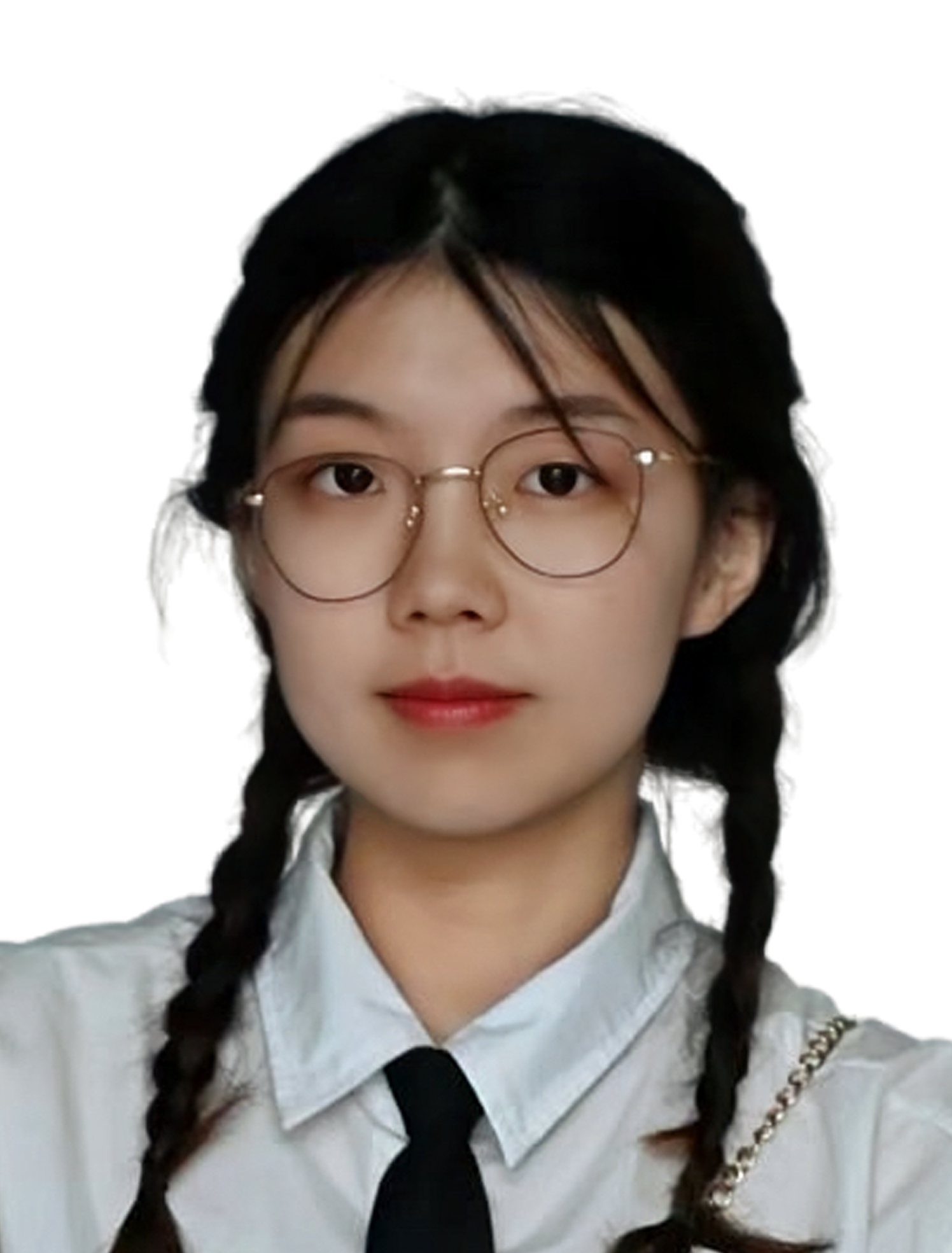}}]
{Can Xu} received the B.Eng. degree in Electrical Engineering from Beijing Forestry University in 2022. She is currently a second-year master’s student in Electrical and Computer Engineering Department at Carnegie Mellon University. Her research interest include robot perception and machine learning.
\end{IEEEbiography}

\begin{IEEEbiography}
[{\includegraphics[width=1in,height=1.25in,clip,keepaspectratio]{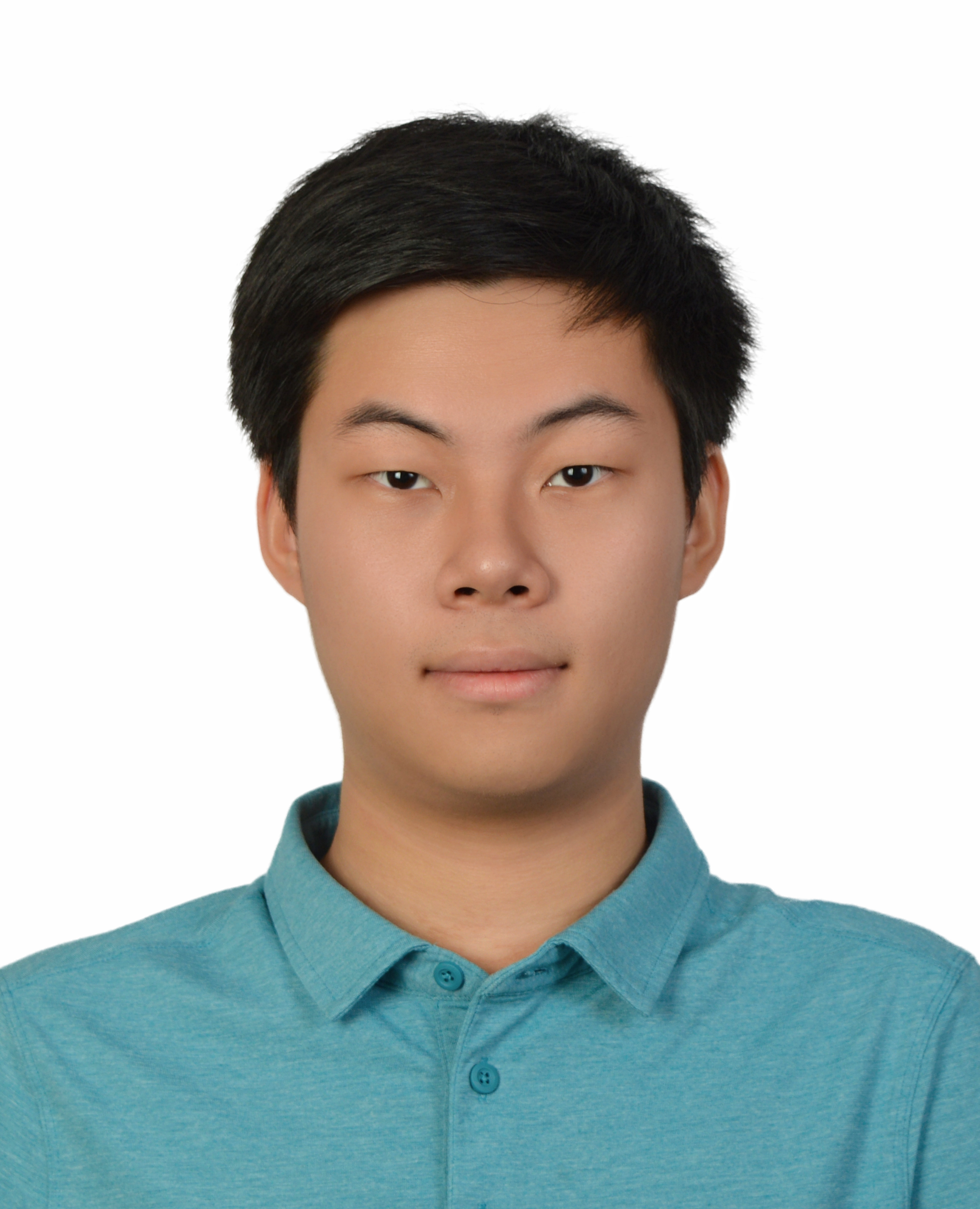}}]
{Yutian Chen} is pursuing a B.Sc. in Computer Science at Carnegie Mellon University with an expected graduation date in 2025. His research interest include robot perception, computer vision and multimodal machine learning.
\end{IEEEbiography}

\begin{IEEEbiography}
[{\includegraphics[width=1in,height=1.25in,clip,keepaspectratio]{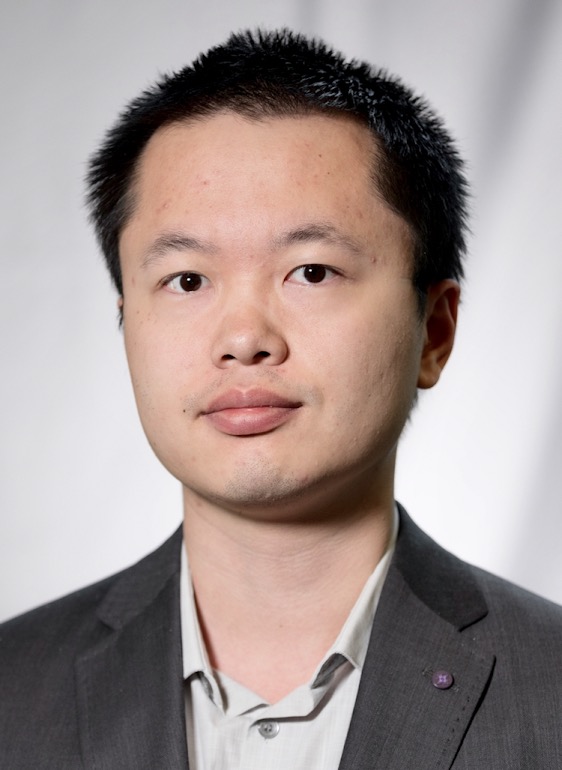}}]
{Xunfei Zhou} obtained the B.Eng. degree from Xi'an Jiaotong University in 2010 and Ph.D. degree in Mechanical Engineering from Texas A\&M University in 2018. Previously, he served as a Staff Research Scientist at OPPO US Research Center. Currently, he works as a Software Engineer at Meta, focusing his research on computer vision, on-device machine learning, and retrieval-augmented generation for multi-modality language models.
\end{IEEEbiography}

\begin{IEEEbiography}
[{\includegraphics[width=1in,height=1.25in,clip,keepaspectratio]{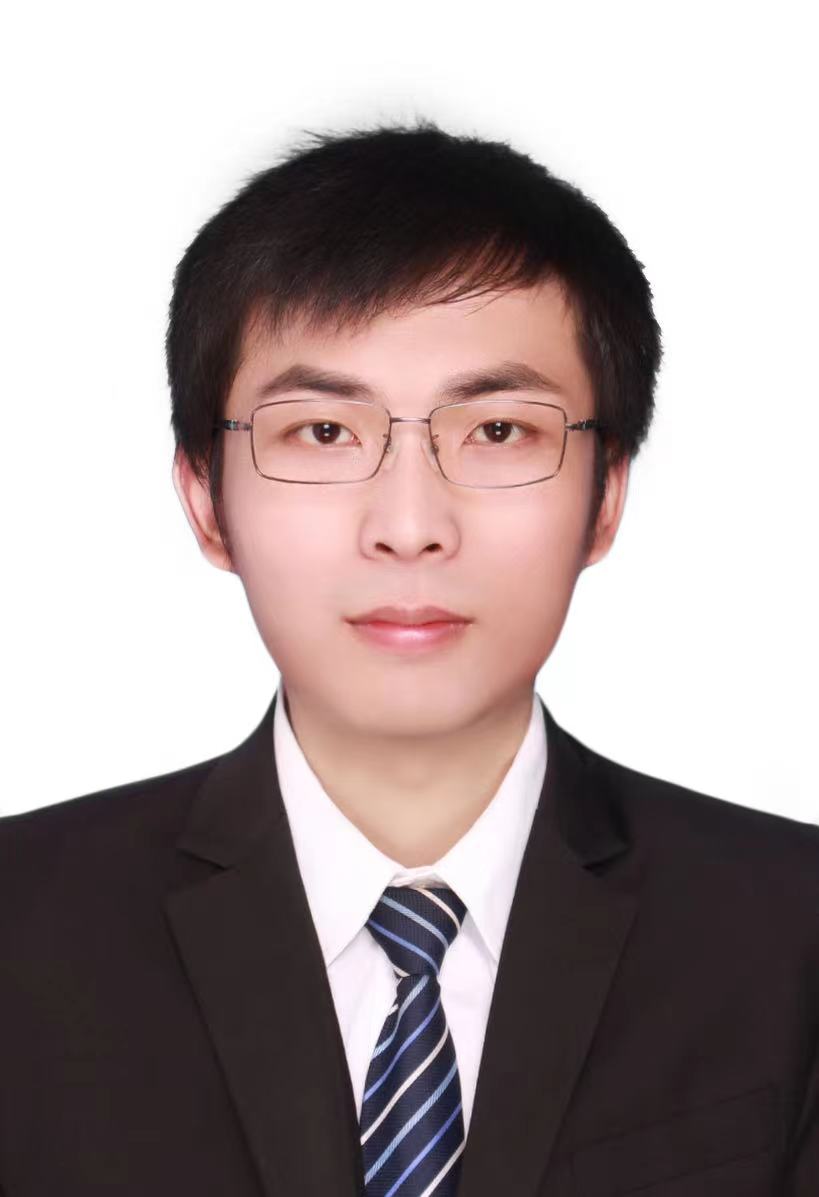}}]
{Youjie Xia}Youjie Xia obtained the M.S. degree from Carnegie Mellon University in 2018. He is currently a Staff Robotics Engineer at OPPO US Research Center. His research interests include machine perception, visual-inertial SLAM, multi-robot localization and mapping. 
\end{IEEEbiography}

\begin{IEEEbiography}[{\includegraphics[width=1in,height=1.25in,clip,keepaspectratio]{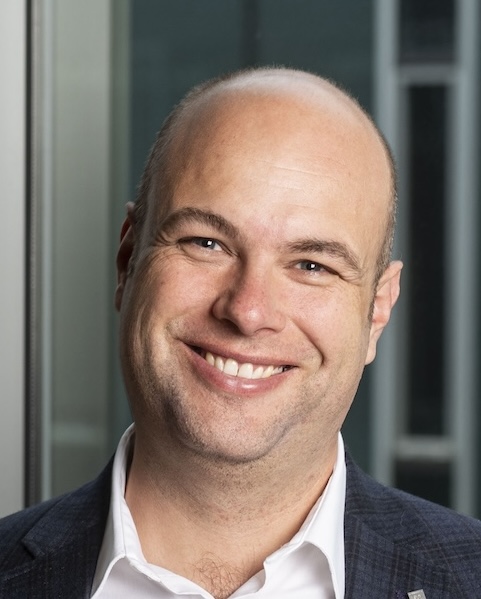}}]
    {Sebastian Scherer}
    received the B.S. degree in computer science and the M.S. and the Ph.D. degrees in robotics from Carnegie Mellon University (CMU), Pittsburgh, PA, USA, in 2004, 2007, and 2010, respectively.

    He is an associate research professor with the Robotics Institute (RI) at CMU. He and his team have shown the fastest and most tested obstacle avoidance on a
    Yamaha RMax (2006), the first obstacle avoidance for microaerial vehicles in natural environments (2008), and the first (2010) and fastest (2014) automatic landing zone detection and landing on a full-size helicopter. His research interest
    includes enabling autonomy for unmanned rotorcraft to operate at low altitude
    in cluttered environments.
\end{IEEEbiography}

}

\end{document}